\documentclass{article}
\usepackage[utf8]{inputenc}
\usepackage{authblk}
\usepackage{setspace}
\usepackage[margin=1.25in]{geometry}
\usepackage{lineno}

\usepackage{epsfig}
\usepackage{subcaption}
\usepackage{calc}
\usepackage{amssymb}
\usepackage{amstext}
\usepackage{amsmath}
\usepackage{amsthm}
\usepackage{multicol}
\usepackage{pslatex}
\usepackage{apalike}

\usepackage{cite}
\usepackage{amsmath,amssymb,amsfonts}
\usepackage{algorithmic}
\usepackage{graphicx}
\usepackage{textcomp}
\usepackage{xcolor}
\usepackage{tabu}
\usepackage{adjustbox}
\usepackage{multirow} 
\usepackage{multicol}
\usepackage{filecontents}
\usepackage{todonotes}
\usepackage{float}
\usepackage{url}
\usepackage{lipsum}

\usepackage{csvsimple}
\usepackage{diagbox}
\usepackage{pgfplots}
\usepackage{subcaption}

\newcommand{\specialcell}[2][c]{%
  \begin{tabular}[#1]{@{}c@{}}#2\end{tabular}}

\usepackage{yhmath}
\usepackage[bb=px]{mathalfa}

\begin{document}

\title{QMRNet: Quality Metric Regression for EO Image Quality Assessment and Super-Resolution}

\author[1*]{David Berga}
\author[2$\dag$]{Pau Gallés}
\author[2$\dag$]{Katalin Takáts}
\author[2$\dag$]{Eva Mohedano}
\author[2$\dag$]{Laura Riordan-Chen}
\author[2$\dag$]{Clara Garcia-Moll}
\author[2$\dag$]{David Vilaseca}
\author[2$\dag$]{Javier Marín}

\affil[1]{Eurecat Centre Tecnològic, Spain.}
\affil[2]{Satellogic Inc, Argentina.}
\affil[*]{Corresponding author. Email: david.berga@eurecat.org}
\affil[$\dag$]{These authors contributed equally to this work.}



\onecolumn \maketitle \normalsize \setcounter{footnote}{0} \vfill

\begin{abstract}

Latest advances in Super-Resolution (SR) have been tested with general purpose images such as faces, landscapes and objects, mainly unused for the task of super-resolving Earth Observation (EO) images. In this research paper, we benchmark state-of-the-art SR algorithms for distinct EO datasets using both Full-Reference and No-Reference Image Quality Assessment (IQA) metrics. We also propose a novel Quality Metric Regression Network (QMRNet) that is able to predict quality (as a No-Reference metric) by training on any property of the image (i.e. its resolution, its distortions...) and also able to optimize SR algorithms for a specific metric objective. This work is part of the implementation of the framework IQUAFLOW\cite{galles2022} which has been developed for evaluating image quality, detection and classification of objects as well as image compression in EO use cases. We integrated our experimentation and tested our QMRNet algorithm on predicting features like blur, sharpness, snr, rer and ground sampling distance (GSD) and obtain validation medRs below 1.0 (out of N=50) and recall rates above 95\%. Overall benchmark shows promising results for LIIF, CAR and MSRN and also the potential use of QMRNet as Loss for optimizing SR predictions. Due to its simplicity, QMRNet could also be used for other use cases and image domains, as its architecture and data processing is fully scalable.

\end{abstract}

\section{\uppercase{Introduction}}
\label{sec:introduction}

One of the main issues in observing and analyzing Earth Observation (EO) images is to estimate its quality. However, this main issue is twofold. Firstly, images are captured with distinct image modifications and distortions, say optical diffractions and aberrations, detector spacings and footprints, atmospheric turbulences, platform vibrations, blurring, target motions, and postprocessing. Secondly, EO image resolution is very limited, due to the sensor's optical resolution, the satellite and connection capacity to send high quality images to ground as well as the captured Ground Sampling Distance (GSD) \cite{Leachtenauer2022}. These limitatons make the image quality assessment (IQA) hard to evaluate for EO particularly, as there are no comparable fine-grained baselines in broad EO domains.

\textbf{
 We will tackle these problems defining a network that acts as a no-reference (blind) metric, assessing quality and optimizing super-resolution on EO images at any scale and modification.
 }
 
\subsubsection*{Contributions}

We briefly summarize below our main contributions:
\begin{itemize}
\item \textbf{We train and validate a novel network (QMRNet) for EO imagery being able to predict any type of based on its quality and distortion }
\item \textbf{(Case 1) We benchmark distinct Super-Resolution Models with QMRNet and compare results with Full-Reference, No-Reference and Feature-based Metrics }
\item \textbf{(Case 2) We benchmark distinct EO datasets with QMRNet scores}
\item \textbf{(Case 3) We propose to use QMRNet as a loss for optimizing quality of super-resolution models}

\end{itemize} 

\subsection{Super-Resolution}

Super-resolution (SR) consists on estimating a high resolution image (HR) given a low resolution one (LR). Initial work from \cite{Zeyde2012} defined a technique named Sparse Coding, which consisted on defining dictionaries of patches specific to the image, which combined were able to reconstruct an HR image. Here the reconstruction error consists on estimating the difference between the reconstructed image (SR) from the LR image and the original HR image and therefore calculate the coefficients for every patch of the dictionary for that particular image in order to reconstruct it.

The first algorithms used for analyzing satellital images were based on multiscale or bilateral filters, only extracting low-level features of the image. Here the denoising problem (similarly what we want to obtain with SR) was tackled by FSRS \cite{Farsiu2004}, with an architecture based on lateral filters trying to minimize variance error with respect the HR. GA-FRSR \cite{Benecki2018} utilized an algorithm able to tune the hyperparameters and kernels from the FRSR. The case of SR-ADE \cite{Zhu2016} utilized a low-level algorithm abse to increase the high-frequency features in order to obtain better image resolution. Another model is the RFSR \cite{Shermeyer2019} utilized 100 random forest regressors with a tree depth of 12. 

Later in the Deep Learning era deep networks have been used to classify images, obtaining high precision on its predictions. For the specific SR task, one can design a network (\textbf{Auto-Encoder}) which its convolutional layers (feature extractor) encode the patches of the image in order to build a feature vector (encoder) from an image, then add deconvolutional layers to reconstruct the original image (decoder). The instances of the predicted images are compared with the original ones in order to re-train the Auto-Encoder network until converging to a HR objective. 

The SRCNN and FSRCNN models \cite{Dong2016,Yamanaka2017} are based on a network of 3 blocks (patch extraction and representation, nonlinear mapping and reconstruction) with 64 filters per layer and 3x3 kernels, using patches of 33x33. The authors also mention to use rotation, scaling and noise transformations as data augmentation prior to training of the network. SRCNN/FSRCNN is said to converge given 100 images using a total of 24.800 patches (crops). The authors use a downscaling using a low-pass filter to obtain LR images and use a bicubic interpolation for the upscaling during reconstruction to obtain the SR image (the predicted HR image). In contrast, for the case of VDSR \cite{Kim_2016_VDSR}, uses a network of 20 convolutional layers (64 filters of 3x3 kernels and ReLU) and a residual map as an additional layer which is able to represent the high-level feature differences between the HR and the interpolated SR image. This residual map is summed to the original LR image in order to obtain the SR image and calculate the reconstruction error and re-train the network. SRCNN has been used by MC-SRCNN \cite{Muller2020} to super-resolve multi-spectral images, by changing the architecture's input channels and adding pan-sharpening filters (modulating the smoothing/sharpening intensity).

These design principles used in autoencoders however have a drawback, that they work differently over feature size frequencies and features at distinct resolutions. For that, multi-scale architectures are proposed. The Multi-Scale Residual Network (MSRN) \cite{Li2018} uses residual connections in multiple residual blocks at different scale bands, non-exclusive to ResNets. It ables to equalize the information bottleneck in deeper layers (high-level features) where the spatial information in some cases tend to diminish or either vanish. Traditional convolutional filters in primary layers have a fixed and narrow field of view, which create dependencies to the learning of spatial long-range connections and deeper layers.  However, multiscale blocks cope with this drawback by analyzing the image domain at different resolution scales to be later merged in a high dimensional multiband latent space. It allow a better abstraction at deeper layers and therefore reconstruct spatial information. This is a remarkable advantage when using EO images, which come with distinct resolutions and GSD.

Novel state-of-the-art SR models are based on \textbf{Generative Adversarial Networks (GANs)}. These networks are composed of two networks, a generator that generates an image estimate (SR), and a discriminator which decides whether the generated image is real or fake under certain categorical or metric objective with respect classification of a set of images "x". Usually, the generator is a deconvolutional network which is fed with a latent vector "z", which represents the distribution for each image. First, the discriminator is trained using an existing database (in order to define the discrimination objective). Then, freezing the discriminator, the generator generates the SR image estimates initializing the latent vector at a random distribution. After that, the discriminator is fine-tuned given the generated SR from the latent space. Finally, the generator is re-trained given the loss obtained from the generator. The main objective of GANs is to maximize the probability of the generator to fool the discriminator. In the SR problem, the LR is considered as the input latent space "z" while the HR image is considered as the real image "x" to obtain the adversarial loss. For the case of the popular SRGAN \cite{Ledig2017}, it has been designed with adversarial loss through VGG and ResNet (SRResnet), with residual connections and perceptual loss. Its generator uses 16 residual blocks, and for each residual block there are 2 convolutional layers of 64 filters and 3x3 kernels, batch normalization and ReLU. The discriminator has 8 convolutional layers followed of 3x3 kernels with filters scaled with a factor of 2, from 64 to 512 filters, followed by fully-connected layers and a sigmoid for the calculation of the adversarial loss. The ESRGAN \cite{Wang2019} is an improved version of the SRGAN although uses adversarial loss relaxation, adds training upon perceptual loss and some residual connections in its architecture.

The main intrinsic difference between GANs and other architectures resides on that the image probability distributions is intrinsically learned. This makes these architectures to suffer from unknown artifacts and hallucinations, however, their SR estimates are usually sharper than autoencoder-type architectures. Some mentioned generative techniques for SR, such as SRGAN/SRResnet, ESRGAN, or Enlighten-GAN \cite{Jiang2021}, and convolutional SR autoencoders, such as VDSR, SRCNN/FRSCNN, or MSRN, don’t adapt their feature generation to optimize a loss based on a specific quality standard that considers all quality properties of the image (both structural and pixel-to-pixel). However, predictions show typical distortions such as blurring (from downscaling the input) or GAN artifacts from the training domain objective. Most of these GAN-based models build the LR inputs of the network from downsampled data from the original HR. This LR generation from downsampling HR limits the training of these models to perform the reverse transformation of the modification, however, the type of distortions and variations from any test image be a combination of much more diverse modifications. The only way to mitigate this limitation, but only partially due to overfitting, is to augment the LR samples to distinct transformations simultaneously.

Some \textbf{self-supervised} techniques can learn to solve the ill-posed inverse problem from the observed measurements, without any knowledge of the underlying distribution assuming its invariance to the transformations. The Content Adaptive Resampler (CAR) \cite{sun2020learned} was proposed, in which a join-learnable downscaling pre-step block together with a upscaling block (SRNet) are trained separately. It is able to learn the downscaling step (through a ResamplerNet) learning the statistics of kernels from the HR image, then it learns the upscaling blocks with another net (SRNet/EDSR) to obtain the SR images. CAR has been able to improve experimental results of SR by considering the intrinsic divergences between LR and HR.

The Local Implicit Image Function (LIIF) \cite{chen2020learning} is able to generate super-resolved pixels considering 2D deep features around these coordinates as inputs. In LIIF, an encoder is jointly trained in a self-supervised super-resolution task maintaining high fidelity in higher resolutions. Since the coordinates are continuous, LIIF can be presented in any arbitrary resolution. Here the main advantage is that the SR is represented in a resolution without resizing HR, making it invariant to the transformations performed to the LR. This ables LIIF to extrapolate SR upon factors up to x30.


\subsection{Image Quality Assessment}

In order to assess the quality of an image, there are distinct strategies. Full-Reference metrics consider the difference between an estimated or modified image (SR) and the reference image (HR). In contrast, no-reference metrics assess the specific statistical properties of the estimated image (SR) without any reference image. Other more novel metrics calculate high-level characteristics of the estimated (SR) image by comparing its distribution distance with respect a preprocessed dataset or either the reference (HR) image in a feature-based space. 

\subsubsection*{Full-Reference Pixel-Level metrics }

The similarity between predicted images (SR) and the reference high-resolution images (HR) is estimated by either looking at the pixel-wise differences responsive to reflectance, sharpness, structure, noise, etc. Very well-known examples of pixel-level metrics are the Root-Mean-Square Error (RMSE) \cite{Pradham2008}, the Peak Signal-to-Noise Ratio (PSNR) \cite{HuynhThu2008}, Structural Similarity Metric (SSIM/MSSIM) \cite{Wang2004}, Haar Perceptual Similarity Index (HAARPSI) \cite{Reisenhofer2018}, Gradient Magnitude Similarity Deviation (GMSD) \cite{Xue2014}, Mean Deviation Similarity Index (MDSI) \cite{ZiaeiNafchi2016}, Spectral Angle Mapper (SAM) \cite{Kruse1993}, Universal Image Quality Index (UQI/UIQ/UIQI) \cite{ZhouWang2002}, Human Visual System-based (HVS) \cite{Sheikh2005}, or Visual Information Fidelity Criterion (IFC/VIF) \cite{Sheikh2006}. 

The RMSE evaluates the absolute pixel error between SR and HR. For the case of PSNR uses an estimate calculating the power of the signal (SR) considering the noise error with respect the HR image. Some metrics such as the SSIM specifically measure the means and covariances locally for each region at a specific size (e.g. 8x8 patches; multi-scale patches for MSSIM) affecting the overall metric score. The GMSD calculates the global variation similarity of gradient based on a local quality map combined with a pooling strategy. Most comparative studies use these metrics to measure the actual SR quality, mostly relying on PSNR, although there is no evidence that these measurements are the best for EO cases, as some of these are not sensitive to local perturbations (i.e. blurring, over-sharpnening) and local changes (i.e. artifacts, hallucinations) to the image. The HAARPSI and IFC/VIF calculate an index based on the difference (absolute or in mutual information) using the sum of a set of wavelet coefficients processed over the SR-HR images. Other cases of metrics combine some of the pinpointed parameters simultaneously. For instance, the MDSI compares jointly the gradient similarity, chromaticity similarity, and deviation pooling. Other cases such as the NQM \cite{DameraVenkata2000} and UQI/ UIQ /UIQI consider luminance, contrast, and structural statistics of the image, or VMAF \cite{Aaron2015} that combines measurements of VIF, detail loss and luminance pixel differences.

\subsubsection*{No-Reference metrics}

Pixel-reference metrics have a main requirement, which is that the ground-truth HR images are needed to assess a specific quality standard. For the case of no-reference (or blind) metrics, no explicit reference is needed. These rely on a parametric characterization of the enhanced signal based on statistic descriptors, usually linked to noise or sharpness, embedded in high-frequency bands. Some examples are variance, entropy (He), or high-end spectrum (FFT). The main popular metric in EO is the Modulation Transfer Function (MTF), which measures impulse responses in the spatial domain and transfer functions in the frequency domain. This varies upon overall local pixel characteristics mostly present on contours, corners and sharp features in general \cite{Lim2018}. Here the MTF is very sensitive to local changes such as aforementioned (e.g. optical diffractions and aberrations, blurring, motions, etc).
 
Other metrics would use statistics from image patches in combination with multivariate filtering methods to propose score indexes for a given predefined image given its geo-referenced parameter standards. Such methods include NIQE \cite{Mittal2013}, PIQE \cite{VenkatanathN2015} and GIQE \cite{Leachtenauer1997}. The latter is considered for official evaluation of NIIRS ratings \footnote{\url{https://irp.fas.org/imint/niirs.htm}}, considering Ground Sampling Distance (GSD), signal-to-noise (SNR) and the relative edge response (RER) in distinct effective focal lengths of EO images \cite{Thurman2008, Kim2008ImagebasedEA, Li2014}. Note that RER measures the Line Spread Function (LSF) which corresponds to the absolute impulse response also computed by the MTF. 

The Relative Edge Response measures the slope in the edge response (transition). The lower the metric, the blurrier the image is. Taking the derivative of normalized Edge Response produces the Line Spread Function (LSF). The LSF is a 1-D representation of the system Point Sparsity Function (PSF). The width of the LSF at half the height is called the Full Width at Half Maximum (FWHM). The Fourier Transform of the LSF produces the Modulation Transfer Function (MTF). MTF is determined across all spatial frequencies, but can be evaluated at a single spatial frequency, such as the Nyquist frequency. The value of the MTF at Nyquist provides a measure of resolvable contrast at the highest ‘alias-free’ spatial frequency.

\subsubsection*{Feature-based (ML) perceptual metrics}

In \cite{Benecki2018}, the authors argued that the conventional IQA evaluation methods are not valid for EO as the degradation functions and operation hardware conditions do not meet operational conditions. From there it was defined the keypoint feature similarity (KFS) \cite{Liu2021}, which measures edge and keypoint detector statistics to extract information concerned with local features. Through advances in DL in that aspect, deeper network representations have been shown to improve perceptual quality on images, although with higher requirements. The existence of hallucinations and artifacts in predicted SR images is due to several factors related to insufficient training data, learning limitations and optimization functions of the network architecture itself or simply because of common overfitting problems. The concept of perceptual similarity is defined by the score reference on these trained features (i.e. the generator or reconstruction network). These metrics compare distances between latent features from the predicted image and the reference image. Some SoTA methods of perceptual similarity include the Learned Perceptual Image Patch Similarity (LPIPS) \cite{Zhang2018}, which measures the feature maps obtained by the n-th convolution after activation (image-reference layer n) and then calculates similarity using the Euclidean distance between the predicted SR model features and the reference image features. Some other metrics such as the Sliced Wasserstein Distance (SWD) \cite{NEURIPS2019_f0935e4c} or the Fréchet Inception distance (FID) \cite{NIPS2016_8a3363ab} assume a non-linear space modelling for the feature representations to compare, and therefore can adapt better with larger variability or lack of samples in the training image domains.

\subsection{EO Datasets and Related Work}

Most non-feature based metrics shown below are fully unsupervised, namely that there are no current models that specifically can assess image quality invariably from the specific modifications made on images, specially for EO cases. The most novel strategy, ProxIQA \cite{Chen2021} tries to evaluate the quality of an image by adapting the underlying distribution of a GAN given a compressed input. This method has shown to improve quality tested on images from compression datasets Kodak, Teknick and NFLX, although results may vary upon trained image distributions, as shown by JPEG2000, VMAFp and HEVC metrics.  


 
Very few studies on SR use EO images obtained from current worldwide satellites such as DigitalGlobe WorldView-4 \footnote{\url{https://earth.esa.int/eogateway/missions/worldview-4}}, SPOT \footnote{\url{https://earth.esa.int/eogateway/missions/spot}}, Sentinel-2 \footnote{\url{https://sentinels.copernicus.eu/web/sentinel/missions/sentinel-2}}, Landsat-8 \footnote{\url{https://www.usgs.gov/landsat-missions/landsat-8}}, Hyperion/EO-1 \footnote{\url{https://www.usgs.gov/centers/eros/science/usgs-eros-archive-earth-observing-one-eo-1-hyperion}}, SkySat \footnote{\url{https://earth.esa.int/eogateway/missions/skysat}}, Planetscope \footnote{\url{https://earth.esa.int/eogateway/missions/planetscope}}, RedEye \footnote{\url{https://space.skyrocket.de/docs_dat/red-eye.htm}}, QuickBird \footnote{\url{https://earth.esa.int/eogateway/missions/quickbird-2}}, CBERS \footnote{\url{https://www.satimagingcorp.com/satellite-sensors/other-satellite-sensors/cbers-2/}}, Himawari-8 \footnote{\url{https://www.data.jma.go.jp/mscweb/data/himawari/}}, DSCOVR EPIC \footnote{\url{https://epic.gsfc.nasa.gov/}} or PRISMA \footnote{\url{https://www.asi.it/en/earth-science/prisma/}}. In our study we selected a variety of subsets (see Table \ref{tab:datasets}) from distinct online General Public Domain satellite imagery datasets with high resolution (around 30 cm/px). Most of these are used for land use classification tasks, with coverage category annotations and some with object segmentation. Inria Aerial Image Labeling Dataset (Inria-AILD) \footnote{\url{https://project.inria.fr/aerialimagelabeling/}} contains 180 train and 180 test images, covering 405+405 $km^2$ of US (Austin, Chicago, Kitsap County,  Bellingham, Bloomington, San Francisco) and Austria (Innsbruck  Eastern/Western Tyrol, Vienna) regions. Inria-AILD was used for semantic segmentation of buildings contest. Some land cover categories are considered for aerial scene classification in DeepGlobe \footnote{\url{http://deepglobe.org/}} (Urban, Agriculture, Rangeland, Forest, Water or Barren); USGS \footnote{\url{https://data.usgs.gov/datacatalog/}} and UCMerced \footnote{\url{http://weegee.vision.ucmerced.edu/datasets/landuse.html}} with 21 classes (i.e. agricultural, airplane, baseballdiamond, beach, buildings, chaparral, denseresidential, forest, freeway, golfcourse, harbor, intersection, mediumresidential, mobilehomepark, overpass, parkinglot, river, runway, sparseresidential, storagetanks and tenniscourt). The latter has been captured on many US regions, i.e. Birmingham, Boston, Buffalo, Columbus, Dallas, Harrisburg, Houston, Jacksonville, Las Vegas, Los Angeles, Miami, Napa, New York, Reno, San Diego, Santa Barbara, Seattle, Tampa, Tucson and Ventura. XView\footnote{\url{http://xviewdataset.org/}} contains 1.400 $km^2$ RGB pan-sharpened images from DigitalGlobe WorldView-3 with 1 million labeled objects and 60 classes (e.g. Building, Hangar, Train, Airplane, Vehicle, Parking Lot) annotated both with bounding boxes and segmentation. Kaggle Shipsnet \footnote{\url{https://www.kaggle.com/datasets/rhammell/ships-in-satellite-imagery}} contains 7 San Francisco Bay harbor images and 4000 individual crops of ships captured in the dataset.


\begin{table}[h]
\centering
\small
\begin{tabular}{|l|l|l|l|l|l|l|}
\hline
 Dataset-Subset & \#Set / \#Total & GSD & Resol & Spatial Coverage & Year & Provider \\
 \hline
 USGS & \underline{279}/279  & 30 cm/px & 5000x5000 & 349 km$^2$ (US regions) & 2000 & USGS (LandSat) \\
 UCMerced-380 & \underline{380}/2100 & 30 cm/px & 256x256 & 1022/5652 (US regions) & 2010 & USGS (LandSat)\\
 UCMerced-2100 & \underline{2100}/2100 & 30 cm/px & 232x232 & 5652 km$^2$ (US regions) & 2010 & USGS (LandSat)\\
 Inria-AILD-10-train & \underline{10}/360 & 30 cm/px & 5000x5000  &  22/810 km$^2$ (US \& Austria) & 2017 & arcGIS \\
 Inria-AILD-180-train & \underline{180}/360 & 30 cm/px & 5000x5000  & 405/810 km$^2$ (US \& Austria) & 2017 & arcGIS \\
 Inria-AILD-180-test & \underline{180}/360 & 30 cm/px & 5000x5000  & 405/810 km$^2$ (US \& Austria) & 2017 & arcGIS \\
 Shipsnet-Scenes & \underline{7}/7  & 3m/px & 3000x1500 & 28 km$^2$ (San Francisco Bay) & 2018 & Open California \\
 Shipsnet-Ships & \underline{4000}/4000  & 3m/px & 80x80 & 28 km$^2$ (San Francisco Bay) & 2018 & (Planetscope) \\ 
 DeepGlobe & \underline{469}/1146 & 31 cm/px & 2448x2448 & 703/1.717 km$^2$ (Germany) & 2018 & Worldview-3 \\
 Xview-train & \underline{846}/1127  & 30 cm/px & 5000x5000 & 1050/1.400 km$^2$ (Global) & 2018 & WorldView-3 \\ 
 Xview-validation & \underline{281}/1127 & 30 cm/px & 5000x5000 & 349/1.400 km$^2$ (Global) & 2018  & WorldView-3 \\ 
 \hline
\end{tabular}
\caption{List of datasets used in our experimentation. We show 11 subsets collected from 7 datasets provided in 4 satellites.}
\label{tab:datasets}
\end{table}

\section{\uppercase{Proposed Method}}

\subsection{IQUAFLOW Modifiers and Metrics}

We have developed a set of modifiers\cite{galles2022} that apply a specific type of distortion in EO images \footnote{\url{https://github.com/satellogic/iquaflow}}. In the modifiers list (see Table \ref{tab:modifiers}) we describe 5 modifiers we developed for our experimentation, 3 of which have been integrated from common libraries (Pytorch \footnote{\url{https://pytorch.org/vision/stable/transforms.html}}, PIL \footnote{\url{https://pillow.readthedocs.io/en/stable/reference/Image.html}}), such as Blur, Sharpness and Resolution (GSD) and 2 which we specifically developed to represent RER and SNR blind metric modifications. For the case of Blur, we build a gaussian filter with kernel 7x7 and we parametrize the $\sigma$. For the case of Sharpness, similarly, we build a function that is modulated by a gaussian factor (similar to a $\sigma$). If the factor is higher than 1.0 (i.e. from 1.0 to 10.0), the image is sharpened (high-pass filter, with negative values on the sides of the kernel). However, if the factor is lower than 1.0 (i.e. from 0.0 to 1.0) then the image is blurred through a gaussian function (low-pass filter with gaussian shape). For the case of GSD (Ground Sampling Distance), we apply a bilinear interpolation on the original image to a specific scaling (e.g. x1.5, x2) which will increase the GSD. In this case, an interpolated version of a 5000x5000 image of GSD 30cm/px will be 10000x10000 and its GSD 60cm/px, as its resolution has changed but the (oversampled / fake) sampling distance is doubled (worse). For the case of RER we get the real RER value from the ground truth and calculate the LSF and max value of edge response. From that we build a gaussian function that is adapted to the expected RER coefficients and then filter the image. For SNR, similarly to RER, we require annotation of base SNR from the original dataset. From that, we build a randomness regime that is adapted to a gaussian shape which will be summed to the original image (adding randomness with a specific $\sigma$ slope probability). 


\begin{table}[h]
\small
\centering
\begin{tabular}{|l|l|l|l|l|l|}
\hline
  Modifier & Acronym & Parameters & \#Intervals (N) & Range & Properties\\
 \hline
 Gaussian Blur & $blur$ & Sigma ($\sigma$) & 50 & .0 to 2.5 & $Quality\downarrow, Distortion\uparrow$\\ 
 Gaussian Sharpness & $F$ & Sharpness Factor ($F$) & 9 & 1.0 to 10.0 & $Quality\downarrow, Distortion\uparrow$\\ 
 Ground Sampling Distance & GSD & GSD or scaling & 10 & \specialcell{.30 to .60\\(x1...x2)} & $Quality\downarrow, Distortion\uparrow$\\
 Relative Edge Response & $rer$ & RER (MTF-Sharpness) & 40 & .15 to .55 & $Quality (GT), Distortion\downarrow$\\
 Signal-to-Noise Ratio & $snr$ & Noise (Gaussian) Ratio & 40 & 15 to 30 & $Quality (GT), Distortion\uparrow$\\
 \hline
\end{tabular}
\caption{List of modifier parameters used in QMRNet. These modify the input images and annotate them to provide train and test data for the QMRNet. Distinct intervals have been selected according to the precision and variability of the modification.}
\label{tab:modifiers}
\end{table}

\begin{table}[h!]
\centering
\begin{tabular}{c|c|c|c|c|} 
 \hfill & original & $\xleftarrow{}$lower & & $\xrightarrow{}$higher \\
 \hline
 \parbox[t]{2mm}{\multirow{3}{*}{\rotatebox[origin=c]{90}{Blur ($\sigma$)}}} & \specialcell{\includegraphics[width=0.205\textwidth]{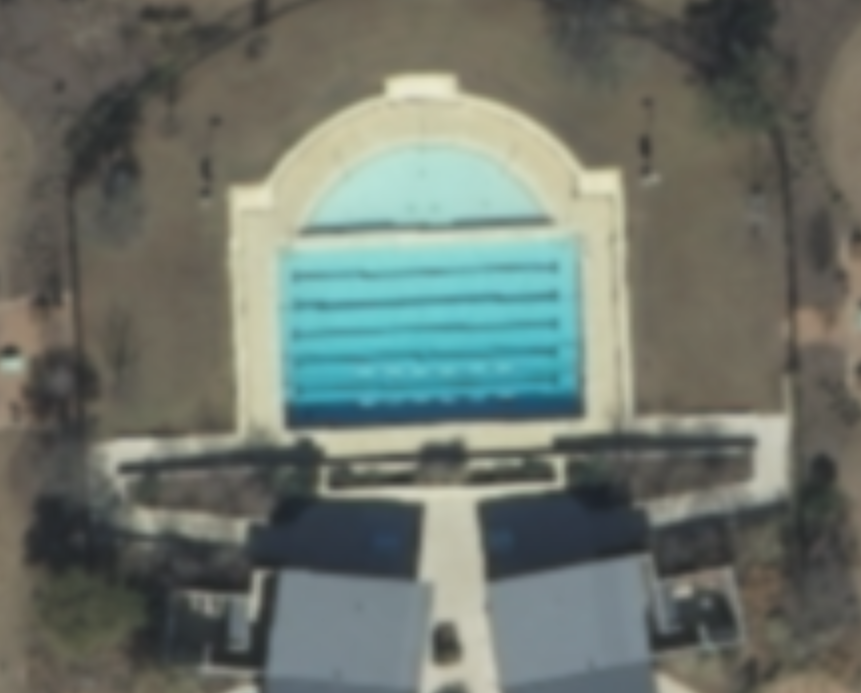}\\$\sigma=1.0$}  & \specialcell{\includegraphics[width=0.205\textwidth]{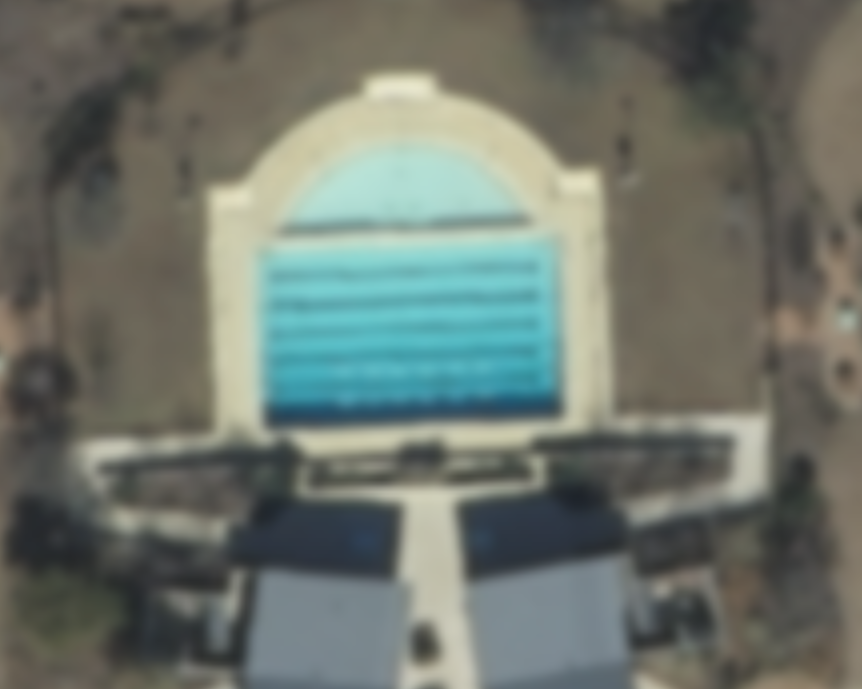}\\$\sigma=1.5$} & \specialcell{\includegraphics[width=0.205\textwidth]{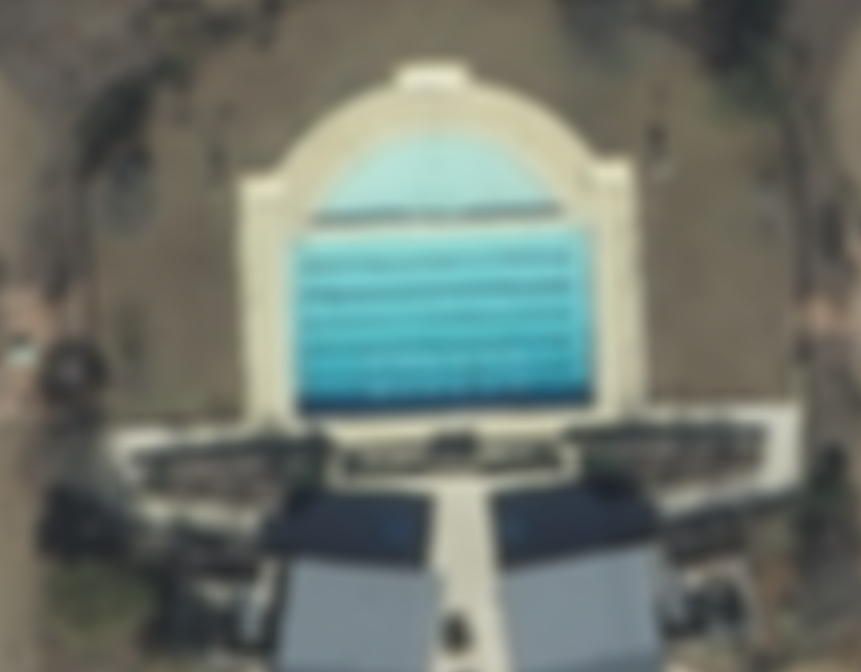}\\$\sigma=2.0$} & \specialcell{\includegraphics[width=0.205\textwidth]{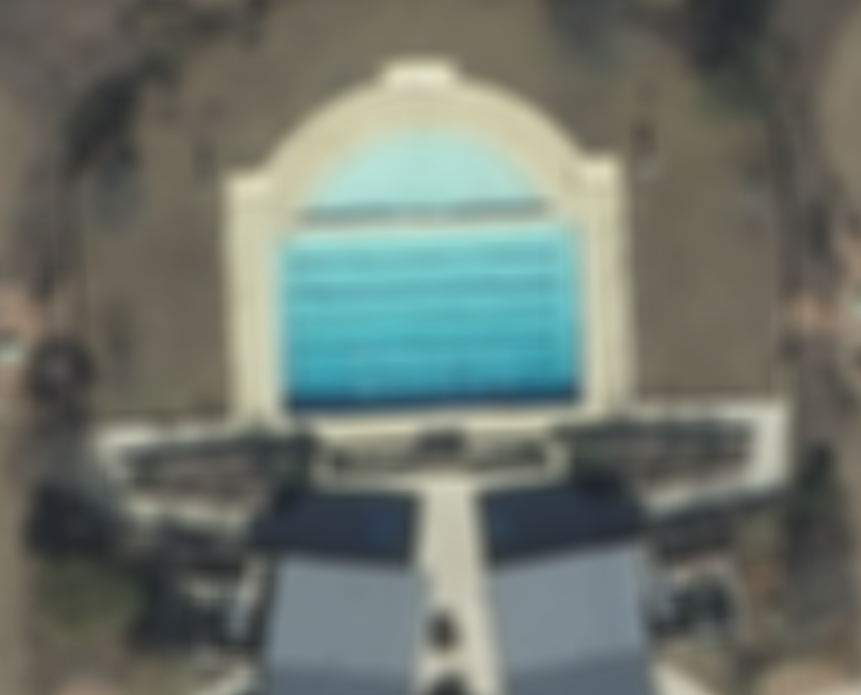}\\$\sigma=2.5$}\\\hline
\parbox[t]{2mm}{\multirow{3}{*}{\rotatebox[origin=c]{90}{Sharpness ($F$)\hspace{-24pt}}}} & \specialcell{\includegraphics[width=0.205\textwidth]{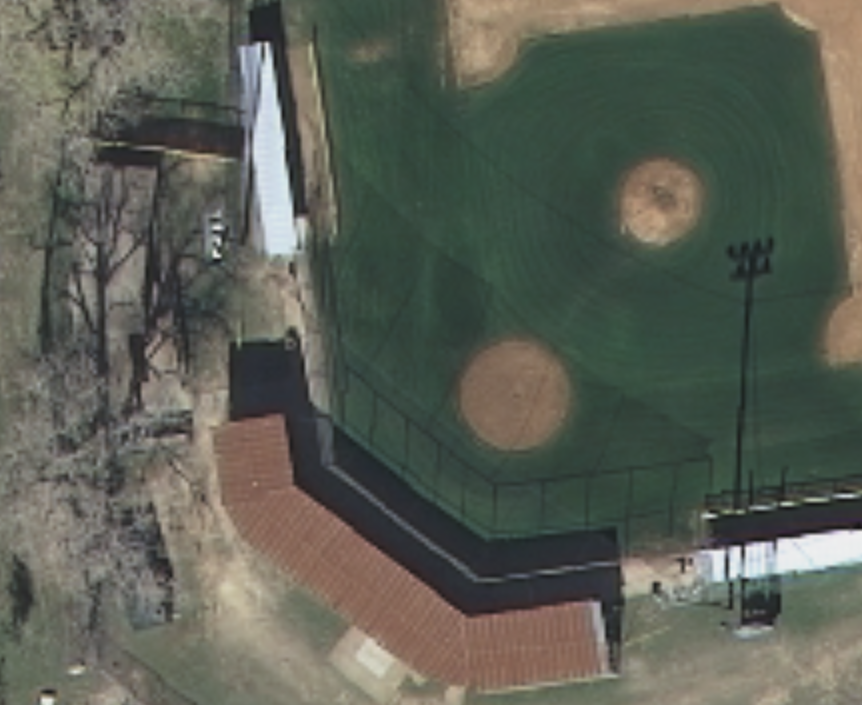}\\$F=1.0$} & \specialcell{\includegraphics[width=0.205\textwidth]{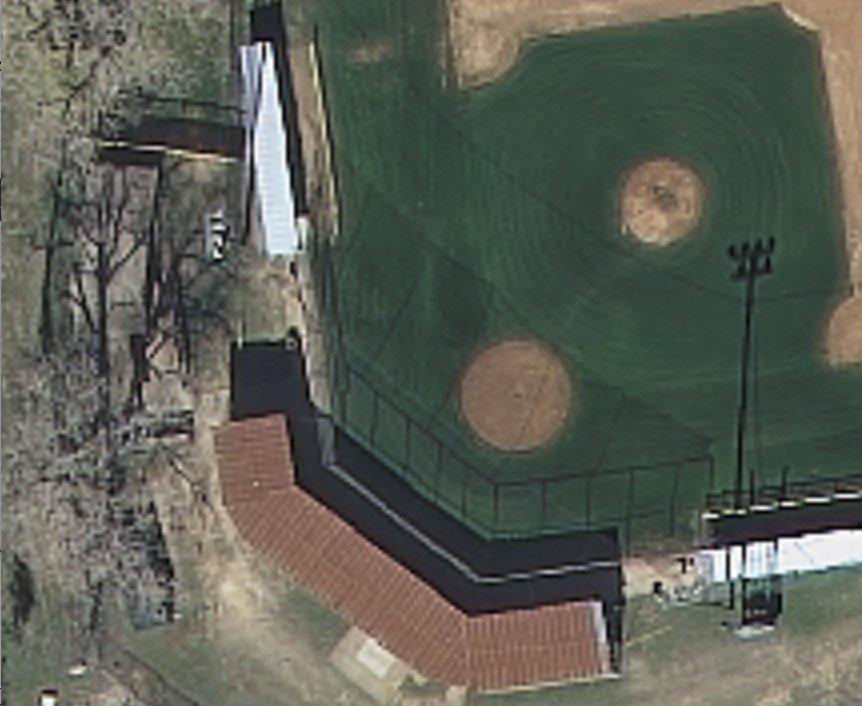}\\$F=2.0$} & \specialcell{\includegraphics[width=0.205\textwidth]{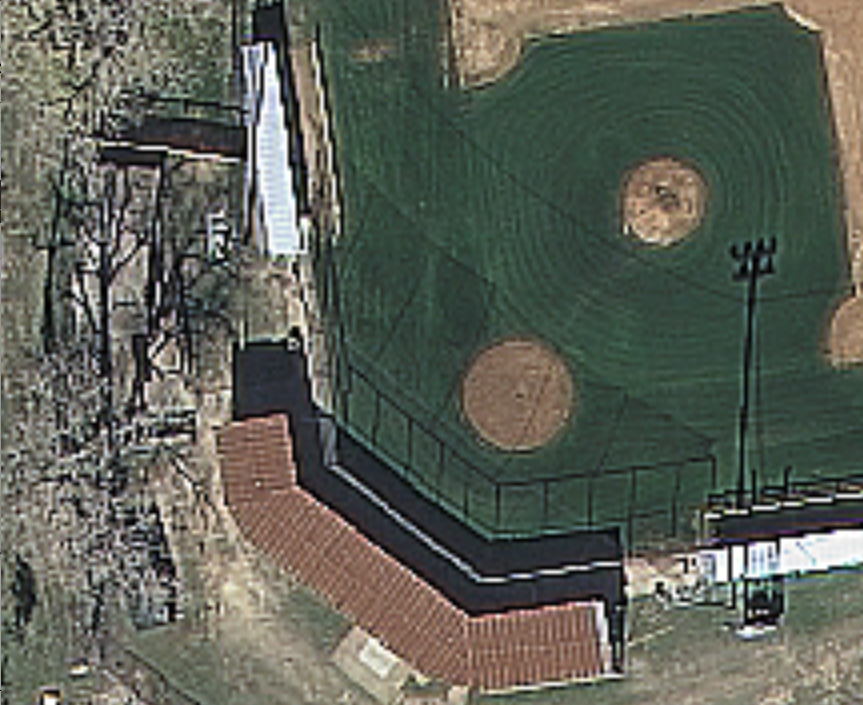}\\$F=5.0$} & \specialcell{\includegraphics[width=0.205\textwidth]{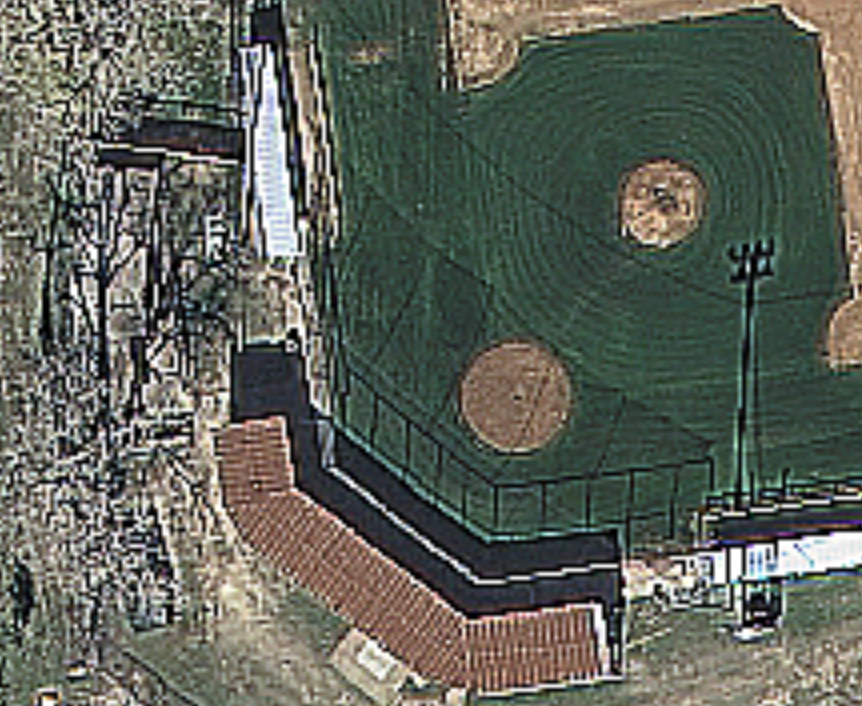}\\$F=10.0$}\\\hline
\parbox[t]{2mm}{\multirow{3}{*}{\rotatebox[origin=c]{90}{GSD}}} & \specialcell{\includegraphics[width=0.205\textwidth]{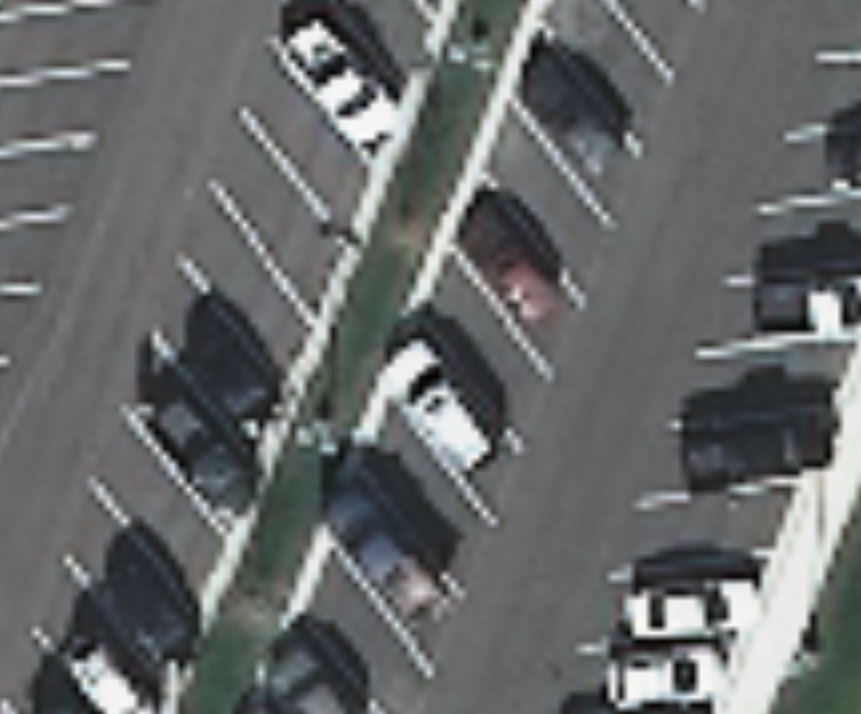}\\$30$\\zoom(x800)} & \specialcell{\includegraphics[width=0.205\textwidth]{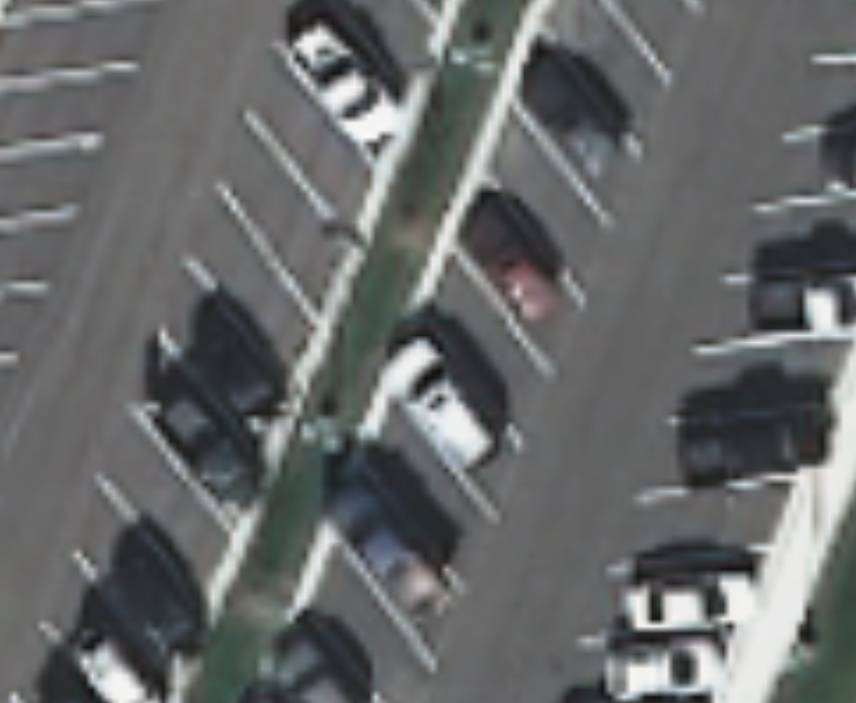}\\$36.\wideparen{6}$\\zoom(x720)} & \specialcell{\includegraphics[width=0.205\textwidth]{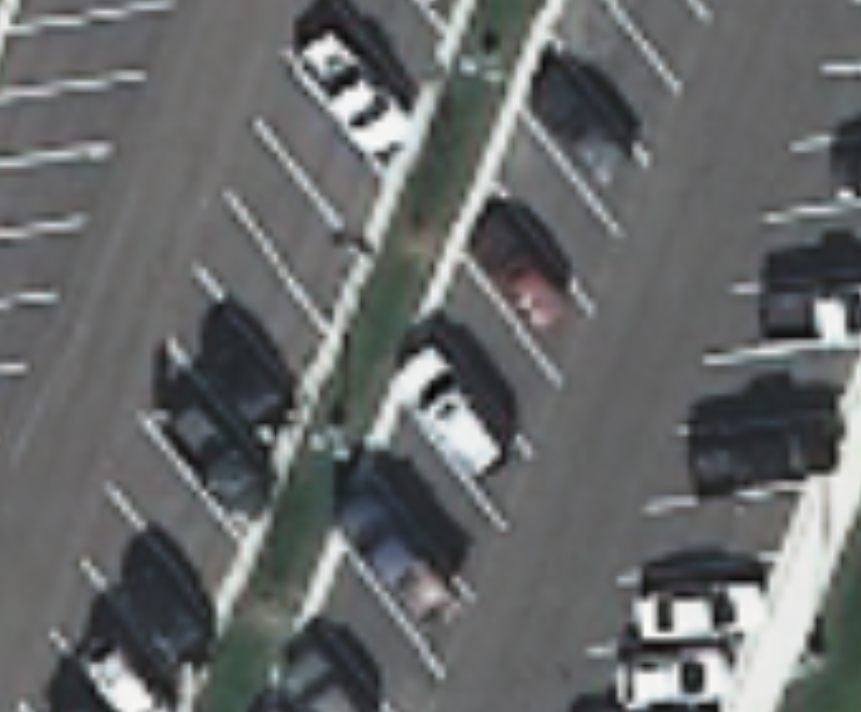}\\$50$\\zoom(x506)} & \specialcell{\includegraphics[width=0.205\textwidth]{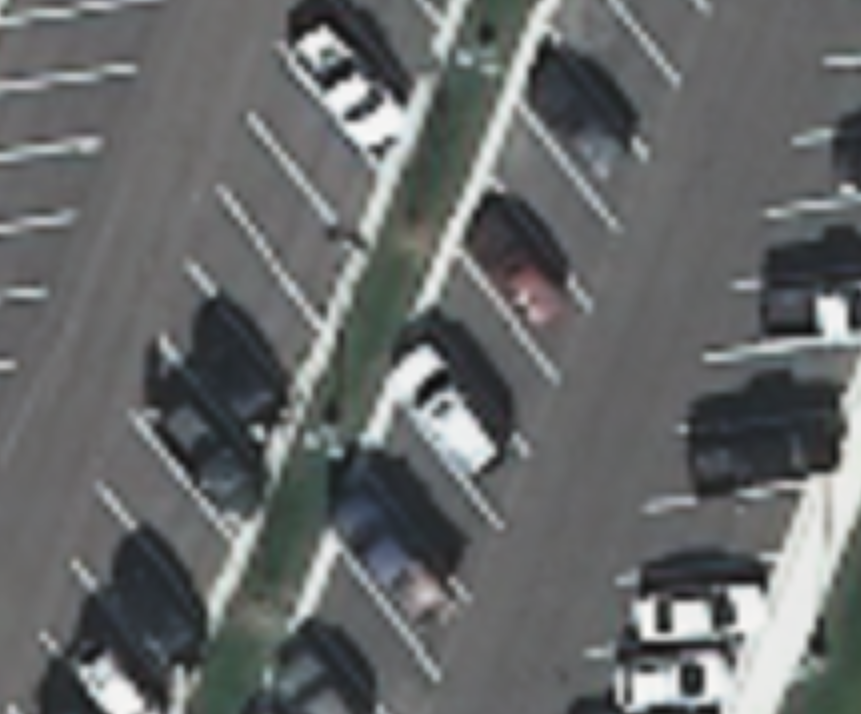}\\$60$\\zoom(x400)}\\\hline
\parbox[t]{2mm}{\multirow{3}{*}{\rotatebox[origin=c]{90}{RER}}} & \specialcell{\includegraphics[width=0.205\textwidth]{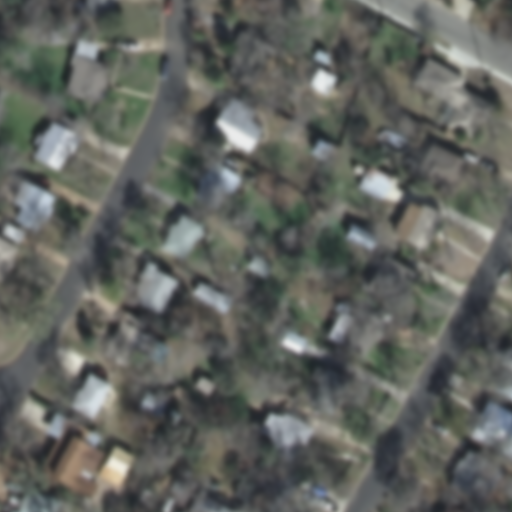}\\$.15$} & \specialcell{\includegraphics[width=0.205\textwidth]{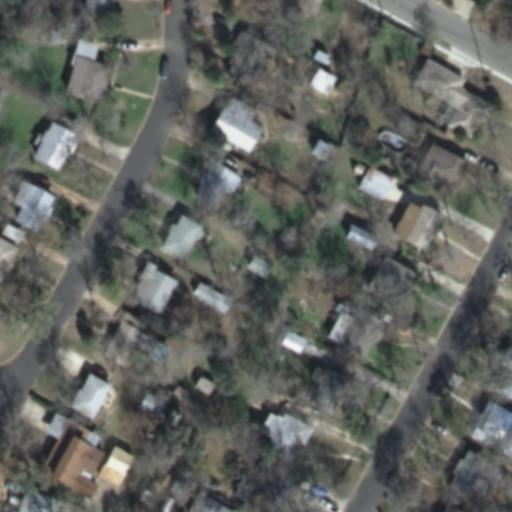}\\$.25$} & \specialcell{\includegraphics[width=0.205\textwidth]{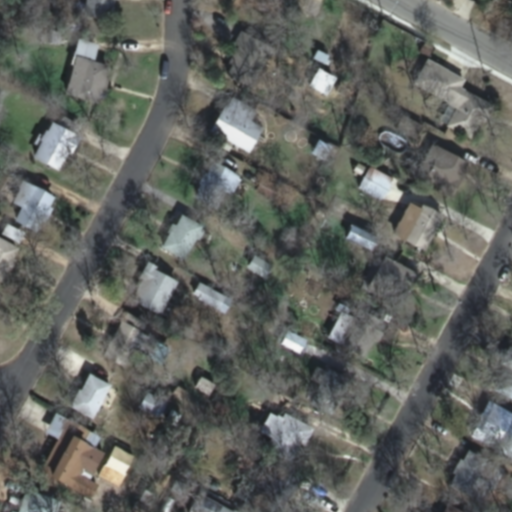}\\$.30$} & \specialcell{\includegraphics[width=0.205\textwidth]{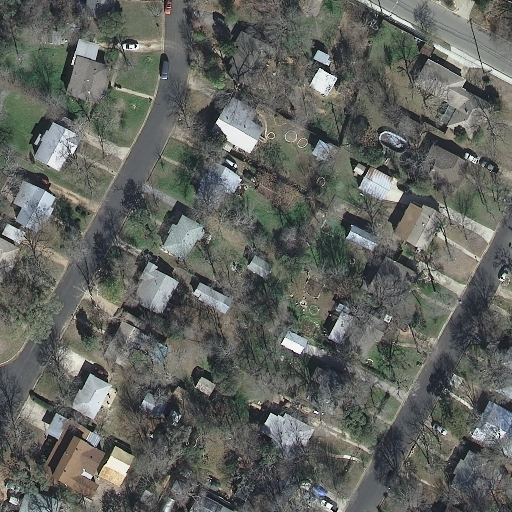}\\$.55$}\\\hline
\parbox[t]{2mm}{\multirow{3}{*}{\rotatebox[origin=c]{90}{SNR}}} & \specialcell{\includegraphics[width=0.205\textwidth]{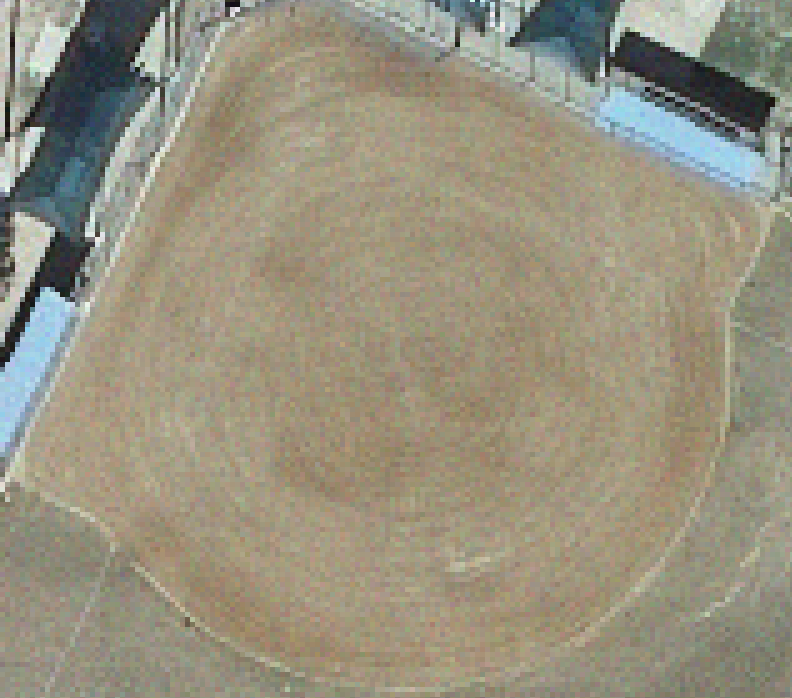}\\$15$} & \specialcell{\includegraphics[width=0.205\textwidth]{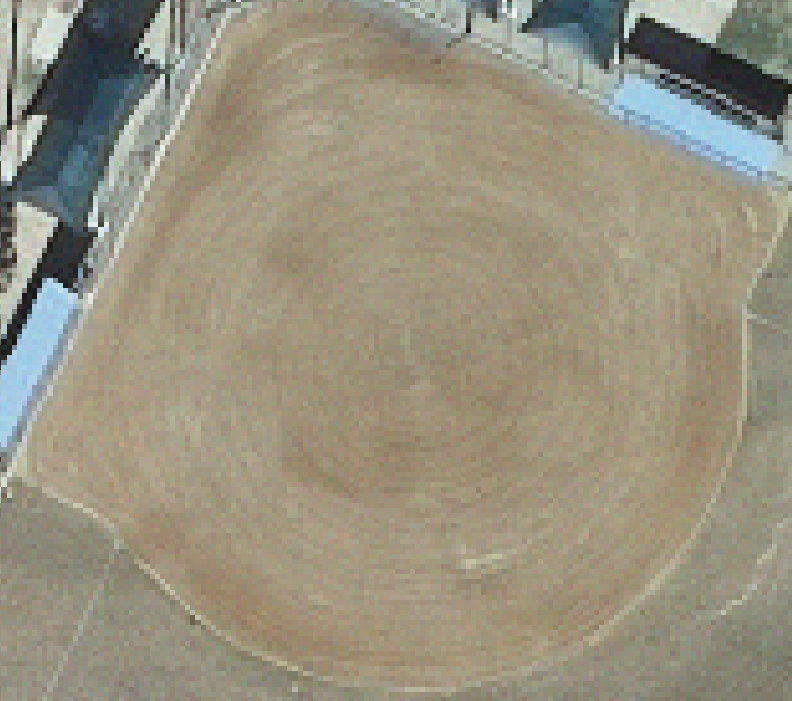}\\$20$} & \specialcell{\includegraphics[width=0.205\textwidth]{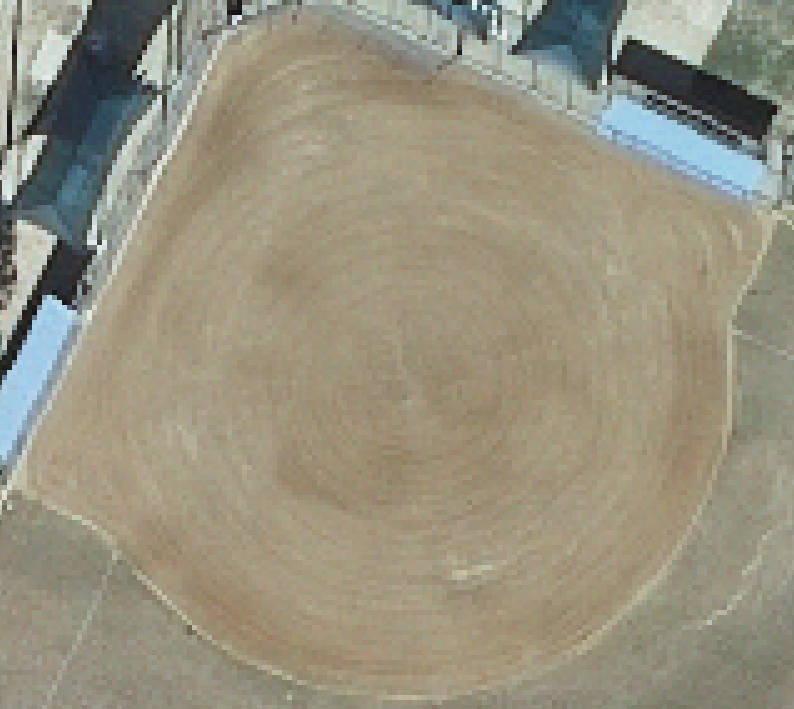}\\$25$} & \specialcell{\includegraphics[width=0.205\textwidth]{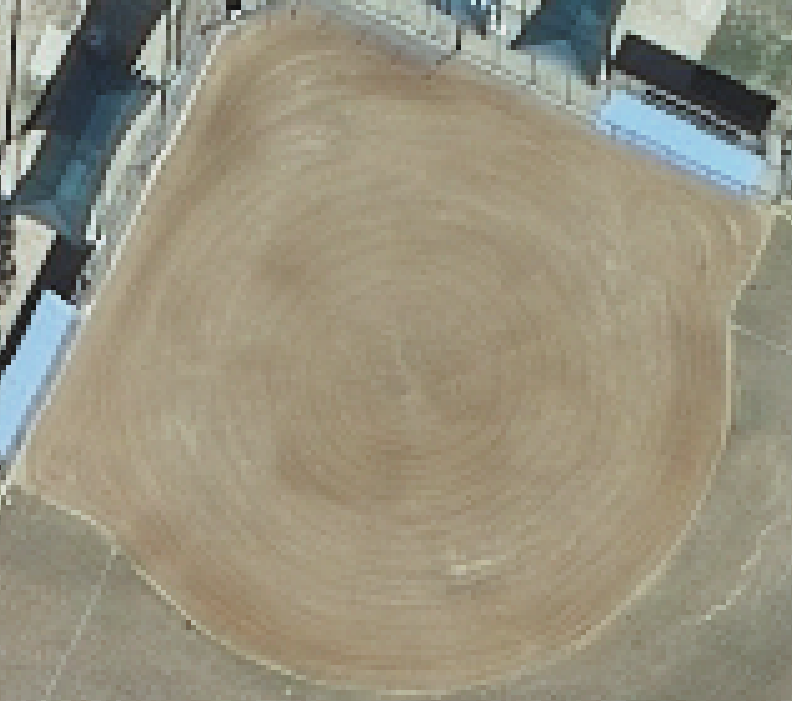}\\$30$}\\\hline
\end{tabular}
\caption{Examples of Inria-AILD crops from modified images for each modifiers (see Table \ref{tab:modifiers})}
\label{tab:examplemodifiers}
\end{table}

\subsection{QMRNet: Classifier of Modifier Metric Parameters}

We have designed the Quality Metric Regression Network (QMRNet) able to regress quality parameters upon the modification or distortion (see Table \ref{tab:modifiers} and Figure \ref{tab:examplemodifiers}) applied to single images. \footnote{Check the code for QMRNet in \url{https://github.com/satellogic/iquaflow/tree/main/iquaflow/quality_metrics}}. Given a set of images, modified through a gaussian blur (sigma),  sharpness (gaussian factor), a rescaling to a distinct GSD, noise (SNR), or any kind of distortion, images are annotated with that parameter. These annotations can be used by training and validating the network upon classifying the intervals corresponding to the annotated parameters.

\begin{figure}[h!]
    \centering
    \begin{subfigure}[c]{0.43\textwidth}
         \centering
         \includegraphics[width=\textwidth]{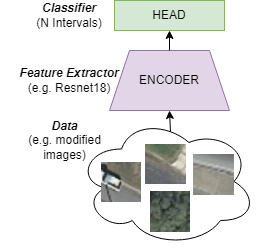}
         \caption{Single parameter QMRNet}
     \end{subfigure}
    \begin{subfigure}[c]{0.43\textwidth}
         \centering
         \includegraphics[width=\textwidth]{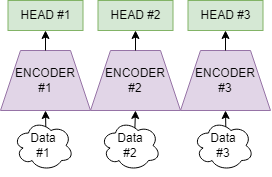}
         \caption{Multiparameter QMRNet-MB}
     \end{subfigure}
    \vspace{5pt}
    \caption{Architecture of QMRNet. (a) Single Parameter QMRNet for using modified / annotated data from a unique parameter / modifier. b) Multi-branch QMRNet (QMRNet-MB), example with 3 stacked QMRNets (3 encoders with 1 head each). } 
    \label{fig:QMRNet}
\end{figure}

QMRNet is a feedfworward neural network that takes architecture of a classifier with a parametizable classifier (see Figure \ref{fig:QMRNet}) upon numerical class intervals (can be set binary, categorical or continuous according to the N intervals). It trains upon predicted interval differences and the annotated parameters of the ground truth (GT), and requires a HEAD for each parameter to predict. We designed 2 mechanisms of assessing quality from several parameters (multiparameter prediction) Multibranch (MB). For MB it is required a single encoder and head for each parameter to predict, while MH requires a head for each parameter but only one encoder. The MH predicts all parameters simultaneously (faster) but its capacity is lower (can lead to lower accuracy) from the encoder part. 

For our experiments with QMRNet we have used an Encoder based on ResNet18 (backbone) composed of a convolutional layer (3x3) and 4 residual blocks (each composed by 4 convolutional layers) of 64, 128, 256 and 512 pixels of resolution. Our network is scalable to distinct crop resolutions as well as regression parameters (N intervals) adapting the HEAD to the number of classes to predict. The output of the HEAD after pooling is a continuous value of probability of each class interval, and through softmax and thresholding we can filter (one-hot) which class or classes have been predicted (1) and not (0) for each image sample crop. By default we utilize the Binary Cross Entropy Loss (BCELoss) as classification error and Stochastic Gradient Descend as optimizer. For the case of Multiclass regression, in QMRNet-MB we train each network individually with its set of parametized modification intervals for each sample. 

Note that for processing irregular or inequivalent crops in our design of the network input, in the case of having the encoder input resolution $R$ lower than the input image crops (e.g. 5000x5000 for GT and 256x256 for the network input), we crop the image to the QMRNet input $R$ by $C$ crops. $C$ the number of crops to generate for each sample (e.g. 10, 20, 50, 100, 200). In the case of the crops being smaller than the encoder backbone input (e.g. 232x232 for the GT and 512x512 for the network input), we apply a circular padding on each border (width and/or height) to obtain a real image that preserves scaling and domain. The total number of hyperparameters to specify to design a specific QMRNet architecture is NxR and can be trained with distinct combination of hyperparameters (NxCxR).

To train the QMRNet's regressor, we select a training set and generate a set of distorted cases, which are parameterizable through our modifiers. The total number of training samples (dataset size) can be calculated by the product of dataset images (I) and NxC (number parameter intervals and crops per sample). We can set distinct possible hyperparameters specific to train and validate, such as number of epochs (e), batch size (bs), learning rate (lr), weight decay (wd), momentum, soft thresholding, etc.

\subsection{QMRLoss: Learning Quality Metric Regression as Loss in SR}

\begin{figure}[h]
    \centering
    \includegraphics[width=6cm]{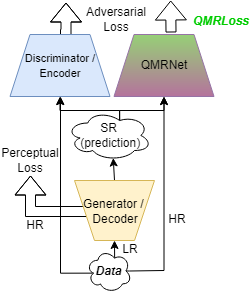}
    \caption{Super-Resolution Model pipeline (Encoder-Decoder for AutoEncoders and Generator-Discriminator for GANs) with ad-hoc QMRNet Loss optimization}
    \label{fig:qmrloss}
\end{figure}

We designed a novel objective function that is able to optimize super-resolution algorithms upon a specific quality objective using QMRNet (see Figure \ref{fig:qmrloss}). 

Given a GAN or AutoEncoder network, we can add an ad-hoc module based on a specific (or several) parameters of QMRNet. The QMRLoss is obtained by computing the classification error between the SR prediction and the original HR image. This classification error determines whether the SR image is distinct in terms of a quality parameter objective (i.e. $blur$, $F$, GSD, $rer$ or $snr$) with respect the HR. The QMRLoss has been designed to use any classification error (i.e. BCE, L1 or L2) and can be summated to the Perceptual or Content Loss of the Generator (Decoder for Autoencoders) in order to tune the SR to the quality objective.

The objective function for image generation algorithms is based on minimizing the Generator (G) error (that compares HR and SR) while maximizing discriminator (D) error (that tests whether the SR image is true or fake).
\begin{equation}
 \min_{G}  \max_{D} = \mathbb{E}_{HR}\left[log(D(I_{HR})\right]+ \mathbb{E}_{LR}\left[log(1-D(I_{SR})\right]
\end{equation}

During training, G is optimized upon $\mathcal{L}^{SR}$, which considers $\mathcal{L}_G^{Perc}$ and $\mathcal{L}^{Adv}$. We added a new term, $\mathcal{L}_G^{QMR}$ which will be our loss function based on quality objectives (QMRNet). Note that here we consider $I_{SR}$ as the prediction image $G(I_{LR})$.
\begin{equation}
 \mathcal{L}^{SR} = \mathcal{L}_G^{Perc} + \mathcal{L}_D^{Adv} + \textcolor{green}{\mathcal{L}_G^{QMR}·\lambda_{QMR}}
\end{equation}

\begin{equation}
\mathcal{L}_D^{Adv} = \sum - log D(I_{SR})
\end{equation}

\begin{equation}
    \mathcal{L}_{G}^{Perc} = \frac{1}{n} \sum (I_{HR} - I_{SR})^{\left[1,2\right]}
\end{equation}

Here we defined the term $\mathcal{L}^{QMR}$, it calculates the parameter difference between HR images and SR images, regularized by the constant $\lambda_{QMR}$. This is done by computing the classification error (L1, L2 or BCE) between the output of the heads for each case:
\begin{equation}
    \begin{split}
    \mathcal{L}^{QMR}_{L1} = \frac{1}{n} \sum (QMRNet(I_{HR}) - QMRNet(I_{SR})) \\    
    \mathcal{L}^{QMR}_{L2} = \frac{1}{n} \sum (QMRNet(I_{HR}) - QMRNet(I_{SR}))^2 \\
    \mathcal{L}^{QMR}_{BCE} = - \frac{1}{n} \sum QMRNet(I_{HR}) · log(QMRNet(I_{SR})) \\ +(1-QMRNet(I_{HR})) · log(1-QMRNet(I_{SR}))
    \end{split}
\end{equation}

\section{\uppercase{Experiments}}
\subsection{Experimental Setup}


For training the QMRNet we collected 30cm/pixel data from INRIA Aerial Image Labeling Dataset (both training and validation using Inria-AILD sets). For testing our network, we selected all the 11 subsets from disctinct EO datasets, USGS, UCMerced, Inria, DeepGlobe, Shipsnet and Xview (see Table \ref{tab:datasets}).

\paragraph{Evaluation Metrics} In order to validate the training regime, we set several evaluation metrics that provide interval-dependencies for each prediction, namely, that intervals that are closer to the target interval are considered better predictions that further ones. This means that given an unblurred image ($blur$ $\sigma=1.0$) the prediction of $\sigma=2.5$ will be a worse prediction than predictions closer to the GT (e.g. $\sigma=1.03$, $\sigma=1.2$). For this, we considered retrieval metrics such as medR or recall rate K (R@K) \cite{Carvalho2018, Salvador2017} as well as performance statistics (Precision, Recall, Accuracy, F-Score) at different intervals close to target (Precision@K, Recall@K, Accuracy@K, F-Score@K) and overall Area Under ROC (AUC). The retrieval metric medR measures the median abolute interval difference between classes, namely, that for 10 classes and modifier GSD ($30$, $33.\wideparen{3}$, $36.\wideparen{6}$, ..., 60), if the targets (modified) are $33.\wideparen{3}$ and predictions are $36.\wideparen{6}$ then there's a medR of 1.0, while if predictions are $60$ then medR is 9.0. R@K measures the total recall (whether prediction in an interval distance from target is lower than K) over a target window (i.e. if there are 40 classes and K is 10, only the 10 classes around the target label are considered for evaluation).

In Tables \ref{tab:results_iqa1} we add another quality metric in addition to the modifier-based ones, which is the $Score$. For this $score$ we defined a basis that describes the overall quality ranking (set from 0.0 to 1.0) of an image or dataset. This is calculated by measuring the weighted mean of metrics, each metric with its own objective target (min$\downarrow$ or max$\uparrow$) as described in Table \ref{tab:modifiers} columns 5-6. 

\begin{equation}
M_{score} = \frac {M_{range} - |(M_{objective}-M_{prediction})}{M_{range}}
\end{equation}
\begin{equation}
Score = \sum_{m=1}^{m=5} \omega^m_M · M^m_{score} 
\end{equation}

For a specific quality metric we define the total $M_{range}$ of the metric (i.e. for $\sigma$ would be 2.5-1, namely, 1.5), an objective $M_{objective}$ value (i.e. for $\sigma$ would be the minimum, as $Quality\downarrow$, namely, 1.0) and an $\omega_M$ which defines the weights for the total weighted sum of the $Score$ (by default if we keep same importance for each metric, $\omega_M=\frac{1}{b}$, where $m$ is the total number of modifiers, for our case $m=5$).

\paragraph{Training and Validation} We trained our network with Inria-AILD sets of 10 and 180 images respectively for short, train and test subsets (Inria-AILD-10, Inria-AILD-train, Inria-AILD-test) with splits of 80/20\%, selecting 100 images for training and 20 for validation (proportionally to 45\% and 12\% from the total respectively). We processed all samples of the dataset with distinct intervals for each modifier (thus, we annotated each sample with that modification interval) and built our network with distinct heads: $N_{blur}=50$, $N_{F}=9$, $N_{GSD}=10$, $N_{rer}=40$, $N_{snr}=40$. We selected a distinct set of crops for each resolution (CxR), in this case 10 crops of 1024x1024, 20 crops of 512x512, 50 crops of 256x256, 100 crops of 128x128 and 200 crops of 64x64. Thus, we generate datasets with different input resolutions but adapting the total domain capacity. The total number of trained images becomes 
 180x$N_{50,9,10,40,40}$x$C_{10,20,50,100,200}$ (e.g. $blur$ 64x64 images set contains 1.8M crop samples).

 We ran our training and validation experiments for 200 epoch with distinct hyperparameters: $lr=[1e-2,1e-3, 1e-4, 1e-5]$, $wd=[1e-3, 1e-4, 1e-5]$, momentum$=0.9$ and soft threshold 0.3 (to filter-out soft to hard/one-hot labels). Due to computational capacity, the training batch sizes have been selected according to the resolution for each set: $bs_{R=64x64, 128x128}=[32,64,128,256]$, $bs_{R=256x256}=[16,32,64,128]$, $bs_{R=512x512}=[8,16,32,64]$ and $bs_{R=1024x1024}=[4,8,16,32]$. 

 In Table \ref{tab:validation_results} we show validation results (Inria-AILD-180-test) for training QMRNet using ResNet18 backbone with Inria-AILD-180-train data. We can observe that the overall medRs are around 1.0 (predictions are about one interval of distance with respect targets) and recall rates (exact match) are for top-1 (R@1) around 70\% and R@5 and R@10 (prediction is in an interval below 5 or 10 of distance with respect target, respectively) around 100\%. This means our network is able to predict parameter data ($blur$ $\sigma$, sharpness $F$, GSD, $snr$, $rer$) with a very high retrieval precision, even when there are 50 classes of intervals. In terms of crop size, best results are mostly in higher input resolution ($R=1024x1024$) .



\begin{table}[h!]
    \centering
    \hspace*{-0.0in}
    \csvautotabular[respect all]{val_inria180.csv}
    \caption{Validation metrics for QMRNet (ResNet18) with all modifiers in Inria-AILD-180-test. *Note that R (Height x Width) defines the resolution input of the network, in each case 1024x1024, 512x512, 256x256, 128x128 and 64x64.} 
    \label{tab:validation_results}
\end{table}

\begin{figure}[H]
\centering
\begin{subfigure}[c]{0.15\linewidth}
\colorbox{green}{\includegraphics[width=\linewidth]{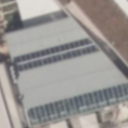}}
\caption*{GT$_\sigma$ = 1.0\\Pred$_\sigma$ = 1.0\\rank(error) = 0}
\end{subfigure} \hspace{0.5pt}
\begin{subfigure}[c]{0.15\linewidth}
\colorbox{green}{\includegraphics[width=\textwidth]{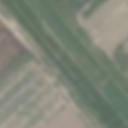}}
\caption*{GT$_\sigma$ = 2.44\\Pred$_\sigma$ = 2.44\\rank(error) = 0}
\end{subfigure} \hspace{0.5pt}
\begin{subfigure}[c]{0.15\linewidth}
\colorbox{red}{\includegraphics[width=\textwidth]{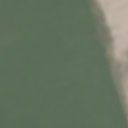}}
\caption*{GT$_\sigma$ = 1.98\\Pred$_\sigma$ = 1.34\\rank(error) = 21}
\end{subfigure} \hspace{0.5pt}
\begin{subfigure}[c]{0.15\linewidth}
\colorbox{green}{\includegraphics[width=\textwidth]{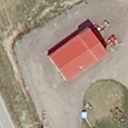}}
\caption*{GT$_F$ = 1.0\\Pred$_F$ = 1.0\\rank(error) = 0}
\end{subfigure} \hspace{0.5pt}
\begin{subfigure}[c]{0.15\linewidth}
\colorbox{green}{\includegraphics[width=\textwidth]{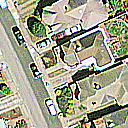}}
\caption*{GT$_F$ = 10.0\\Pred$_F$ = 10.0\\rank(error) = 0}
\end{subfigure} \hspace{0.5pt}
\begin{subfigure}[c]{0.15\linewidth}
\colorbox{red}{\includegraphics[width=\textwidth]{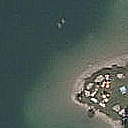}}
\caption*{GT$_F$ = 10.0\\Pred$_F$ = 3.0\\rank(error) = 6}
\end{subfigure}\\
\begin{subfigure}[c]{0.15\linewidth}
\colorbox{green}{\includegraphics[width=\textwidth]{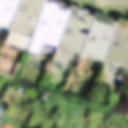}}
\caption*{GT$_{rer}$ = .20\\Pred$_{rer}$ = .20\\rank(error) = 0}
\end{subfigure} \hspace{0.5pt}
\begin{subfigure}[c]{0.15\linewidth}
\colorbox{green}{\includegraphics[width=\textwidth]{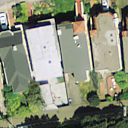}}
\caption*{GT$_{rer}$ = .54\\Pred$_{rer}$ = .54\\rank(error) = 0}
\end{subfigure} \hspace{0.5pt}
\begin{subfigure}[c]{0.15\linewidth}
\colorbox{red}{\includegraphics[width=\textwidth]{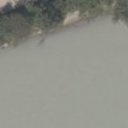}}
\caption*{GT$_{rer}$ = .55\\Pred$_{rer}$ = .37\\rank(error) = 18}
\end{subfigure} \hspace{0.5pt}
\begin{subfigure}[c]{0.15\linewidth}
\colorbox{green}{\includegraphics[width=\textwidth]{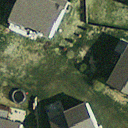}}
\caption*{GT$_{snr}$ = 15\\Pred$_{snr}$ = 15\\rank(error) = 0}
\end{subfigure} \hspace{0.5pt}
\begin{subfigure}[c]{0.15\linewidth}
\colorbox{green}{\includegraphics[width=\textwidth]{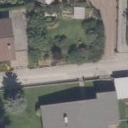}}
\caption*{GT$_{snr}$ = 30\\Pred$_{snr}$ = 30\\rank(error) = 0}
\end{subfigure} \hspace{0.5pt}
\begin{subfigure}[c]{0.15\linewidth}
\colorbox{red}{\includegraphics[width=\textwidth]{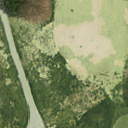}}
\caption*{GT$_{snr}$ = 15\\Pred$_{snr}$ = 30\\rank(error) = 15}
\end{subfigure}\\
\caption{\textcolor{green}{Correct} and \textcolor{red}{Incorrect} prediction examples of QMRNet on Inria-AILD-180 validation (crops resolution, i.e. R = 128x128) given interval rank error (classification label distance between GT and prediction, maximum is N for each net, i.e. 50 for blur, 10 for sharpness and 40 for snr and rer).}
\label{fig:best_worst}
\end{figure}

In Figure \ref{fig:best_worst} we can see that most worst predictions for blur, sharpness, rer and snr appear mainly when attempting to predict over crops with sparse features, namely, when most part of the image has limited or few pixel information (i.e. with similar pixel values), such as sea or flat terrain surfaces. This is because preprocessed samples have few or no dissimilarities in each modifier parameter. This has an effect on evaluating datasets, when surfaces are more sparse, predictions get harder.


\subsection{Results on QMRNet for IQA: Benchmarking Image Datasets} \label{sec:iqa}

We ran our QMRNet with ResNet18 backbone over all the sets described in Table \ref{tab:datasets} \footnote{Check the EO dataset evaluation use case in \url{https://github.com/dberga/iquaflow-qmr-eo}}. Given our network trained uniquely on Inria-AILD-180-train, we see how our network is able to adapt due to the prediction of feasible quality metrics ($blur$ $\sigma$, GSD, sharpness $F$, $snr$ and $rer$) over each of the distinct datasets. We see that with finetuning QMRNet over Inria-AILD-180-train, overall $blur$ for most of the datasets appears to be $\sigma=1.0$ (originally unblurred from the ground truth) except USGS279 and Inria-AILD-test180, which is around $\sigma=1.02$. For the case of sharpness factor $F$, overall values for most datasets is $F=1.0$ (unblurred $\&$ unsharpened) but for cases such as UCMerced380 and Shipnset, appear to be oversharpened ($F>1.5$ and $F>3.0$ respectively). Most datasets present an overall predicted $snr$ of $M(snr)=28,67$ and $rer$ of $M(rer)=.4896$. The highest $Score$ datasets are Inria-AILD-180-train, UCMerced2100 and USGS279, here considering the same weight $\omega_M$ for each modifier metric $M$. 



\begin{table}[H]
    \centering
    \small
    \hspace*{-0.2in}
    \csvautotabular[respect all]{iqa_whole.csv}
    \caption{Mean IQA results of Datasets given QMRNet(ResNet18) trained over 180 images (Inria-AILD-180-train) and 5 modifiers.}
    \label{tab:results_iqa1}
\end{table}

\subsection{Results on QMRNet for IQA: Benchmarking Image Super-Resolution}

Here we selected a set of Super-Resolution algorithms that have been previously tested to super-resolve high quality real image SR dataset benchmarks such as BSD100, Urban100 or Set5x2 \footnote{\url{https://paperswithcode.com/task/image-super-resolution}} but here we want to apply them with EO data and metrics. For this we want to benchmark their performance considering Full-Reference, No-Reference and our QMRNet-based metric
\footnote{Check the use case of Super-Resolution benchmark in \url{https://github.com/dberga/iquaflow-qmr-sisr}}. QMRNet allows us to check the amount of each distortion for every transformation (LR) done to the original image (HR), either if it is the usual x2, x3, x4 downsampling or either a specific distorion such as blurring. 

Concretely, we tested our UCMerced subset of 380 images with crops of 256x256 with AutoEncoder algorithms (FSRCNN and MSRN) and GAN-based and self-supervised architectures such as SRGAN, ESRGAN, CAR and LIIF. All model checkpoints are selected as vanilla (default hyperparameter settings) except for input scaling (x2,x3,x4) and also for the case of MSRN, which we computed the vanilla (untrained) $MSRN_1$ and two other with finetuning $MSRN_2$ (on inria-aid-train180, architecture with 4 scales), and finetuning with added noise $MSRN_3$.

\begin{table}[h!]
    \centering
    \scriptsize
    \hspace*{-.25in}
    \begin{minipage}{.48\linewidth}
    \rotatebox[origin=c]{90}{x2}
    \csvautotabular[respect all]{qmr_x2_blurfalse.csv}
    \end{minipage}
    \hspace*{.25in}
    \begin{minipage}{.48\linewidth}
    \rotatebox[origin=c]{90}{x2+blur}
    \csvautotabular[respect all]{qmr_x2_blurtrue.csv}
    \end{minipage} \\ 
    \hspace*{-.25in}
    \begin{minipage}{.48\linewidth}
    \rotatebox[origin=c]{90}{x3}
    \csvautotabular[respect all]{qmr_x3_blurfalse.csv}
    \end{minipage} 
    \hspace*{.25in}
    \begin{minipage}{.48\linewidth}
    \rotatebox[origin=c]{90}{x3+blur}
    \csvautotabular[respect all]{qmr_x3_blurtrue.csv}
    \end{minipage} \\
    \hspace*{-.25in}
    \begin{minipage}{.48\linewidth}
    \rotatebox[origin=c]{90}{x4}
    \csvautotabular[respect all]{qmr_x4_blurfalse.csv}
    \end{minipage} 
    \hspace*{.25in}
    \begin{minipage}{.48\linewidth}
    \rotatebox[origin=c]{90}{x4+blur}
    \csvautotabular[respect all]{qmr_x4_blurtrue.csv}
    \end{minipage} 
    \caption{Mean No-Reference Quality Metric Regression (QMRNet trained on Inria-AILD-180) Metrics on Super-Resolution of downsampled inputs in UCMerced-380}
    \label{tab:sisr_qmr}
\end{table}

In Table \ref{tab:sisr_qmr} we have evaluated each type of modifier parameter for every single Super-Resolution Algorithm as well as the overall score for all quality metric regression. Here we tested the algorithms considering x2, x3 and x4 downsampling input, as well as considering the case of adding a blur filter with a fixed sigma. Here the QMRNet is able to predict that LR gives the worst ranking for most metrics. FSRCNN and SRGAN give similar results in most metrics, being SRGAN slightly better in blur and sharpness metrics. MSRN shows best results in SNR in most resoltution cases, similarly to SRGAN. For overall scores, CAR presents best results in most metrics, with the highest Score ranking in most downsampling cases. However, CAR has worst ranking in noise metrics, as we mentioned earlier presents oversharpening and hallucidations, which can trick some metrics that measure blur but gets worse for those that predict signal-to-noise ratios. In contrast, LIIF presents worst results in blur and rer metrics, as it appears to be slightly blurred, but has overall good metrics for the rest of modifiers.

In Table \ref{tab:sisr_fullreference} we show a benchmark of known Full-Reference metrics. In super-resolving x2, MSRN (concretely, $MSRN_2$ and $MSRN_3$) gets best results for Full-Reference metrics, including SSIM, PSNR, SWD, FID, MSSIM, HAARPSI and MDSI. In x3 and x4, LIIF and CAR get best results for most Full-Reference metrics, including PSNR, FID, GMSD and MDSI, being top-3 with most metric evaluation. Here we have to pinpoint LIIF do not perform as well when the input (LR) has been blurred, see here that CAR is able to deblur the input better than other algorithms. In Table \ref{tab:sisr_snr} we show No-Reference Metric results, here for SNR, RER, MTF and FWHM. SRGAN, MSRN and LIIF present significantly better results for SNR than other algorithms. This means these algorithms in general do not add noise to the input, namely, the generated images do not contain artifacts that were not present in the original HR image. In this case, CAR gets wost results for SNR but gets best in RER, MTF and FWHM.

\begin{figure}[H]
\centering
\begin{subfigure}[c]{\linewidth}
\begin{tabular}{llllllllll}
    \textit{LR} & \hspace{13pt} FSRCNN & \hspace{0pt} SRGAN & $MSRN_1$ & $MSRN_2$ & $MSRN_3$ &  ESRGAN & CAR & LIIF & \hspace{18pt} \textbf{HR}
\end{tabular}
\end{subfigure}
\begin{subfigure}[c]{\linewidth}
\includegraphics[width=\textwidth]{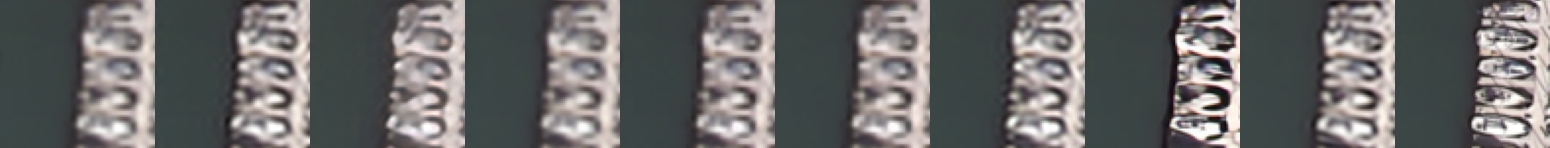}
\end{subfigure}
\begin{subfigure}[c]{\linewidth}
\includegraphics[width=\textwidth]{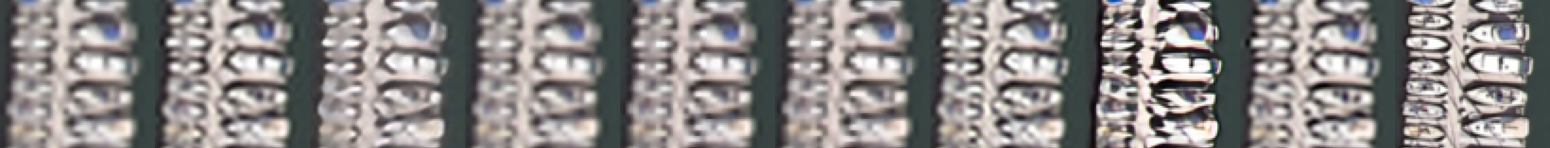}
\end{subfigure}
\begin{subfigure}[c]{\linewidth}
\includegraphics[width=\textwidth]{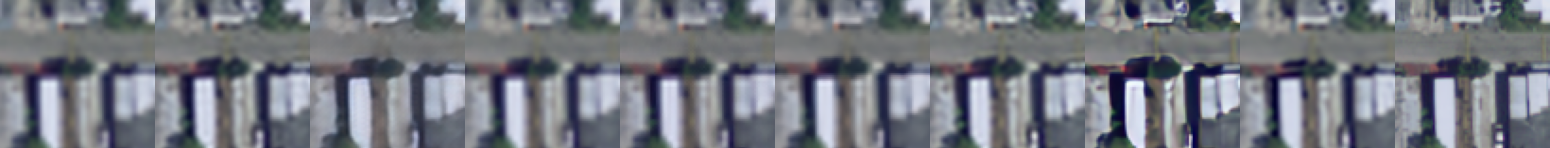}
\end{subfigure}
\begin{subfigure}[c]{\linewidth}
\includegraphics[width=\textwidth]{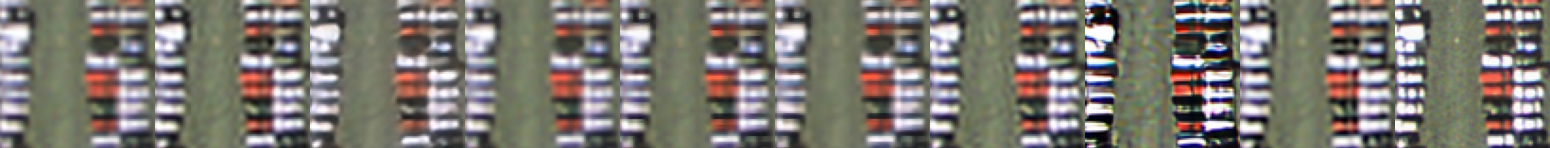}
\end{subfigure}
\begin{subfigure}[c]{\linewidth}
\includegraphics[width=\textwidth]{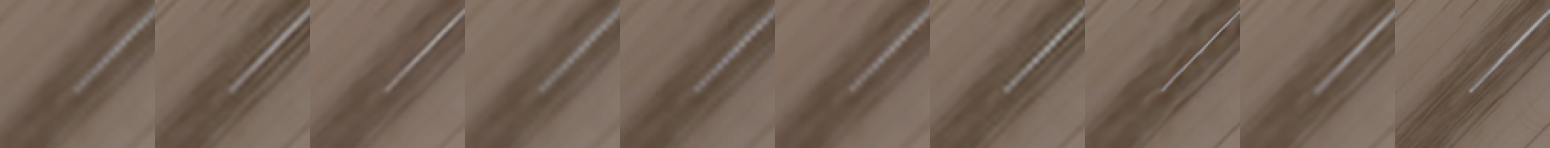}
\end{subfigure}
\begin{subfigure}[c]{\linewidth}
\includegraphics[width=\textwidth]{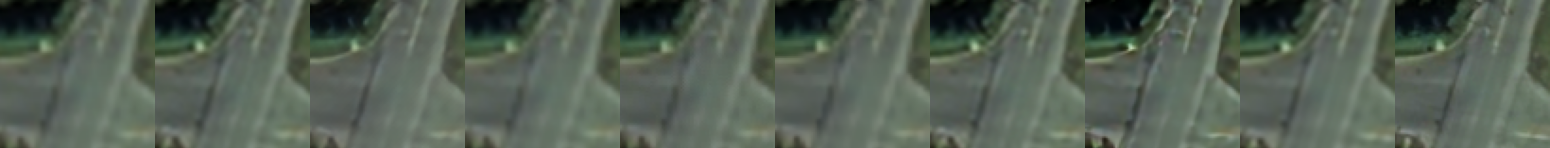}
\end{subfigure}
\begin{subfigure}[c]{\linewidth}
\includegraphics[width=\textwidth]{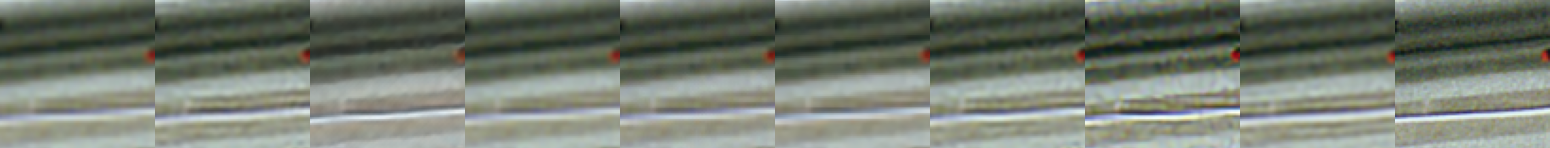}
\end{subfigure}
\begin{subfigure}[c]{\linewidth}
\includegraphics[width=\textwidth]{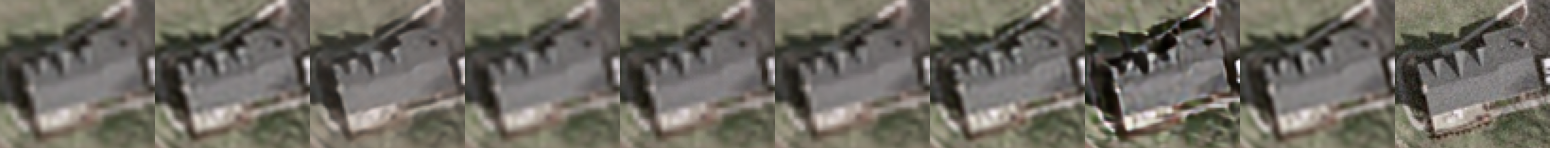}
\end{subfigure}
\begin{subfigure}[c]{\linewidth}
\includegraphics[width=\textwidth]{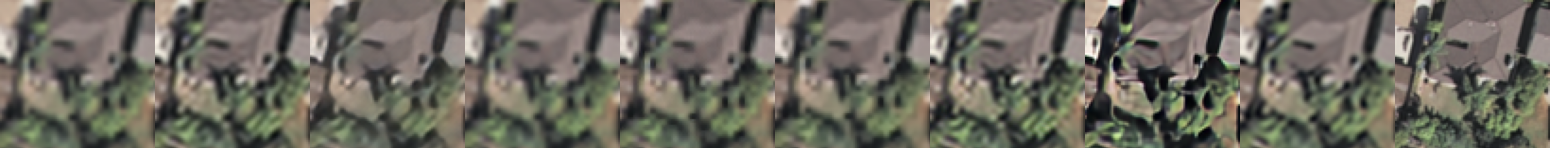}
\end{subfigure}
\begin{subfigure}[c]{\linewidth}
\includegraphics[width=\textwidth]{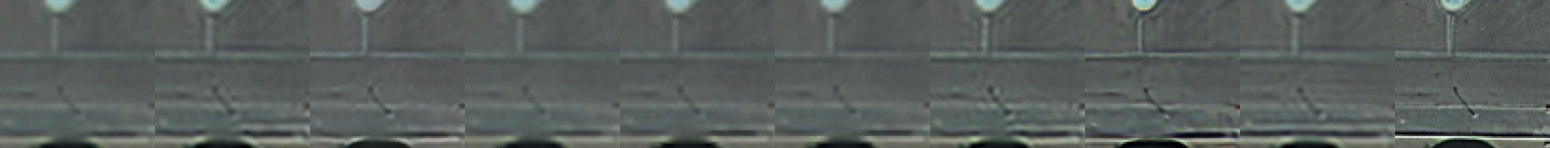}
\end{subfigure}
\begin{subfigure}[c]{\linewidth}
\includegraphics[width=\textwidth]{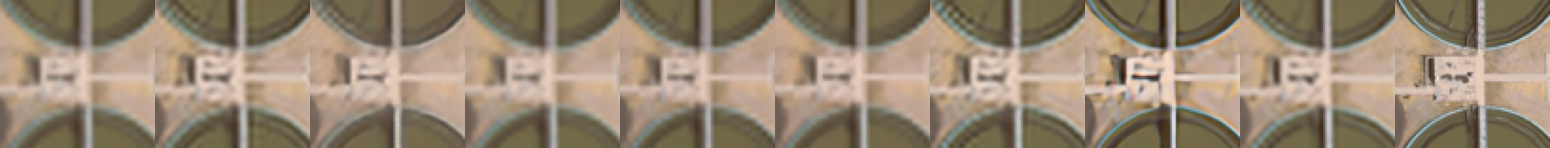}
\end{subfigure}
\caption{Examples (buildings and roads) of UCMerced images (crops of 256x256, with a zoom of x.40 i.e. 100x100) and each SR Algorithm output. \textit{LR} (corresponding to input on algorithms) is the downsampling of \textbf{HR} x3.}
\label{fig:srimages1}
\end{figure}

\begin{figure}[H]
\centering
\begin{subfigure}[c]{\linewidth}
\begin{tabular}{llllllllll}
    \textit{LR} & \hspace{13pt} FSRCNN & \hspace{0pt} SRGAN & $MSRN_1$ & $MSRN_2$ & $MSRN_3$ &  ESRGAN & CAR & LIIF & \hspace{18pt} \textbf{HR} 
\end{tabular}
\end{subfigure}
\begin{subfigure}[c]{\linewidth}
\includegraphics[width=\textwidth]{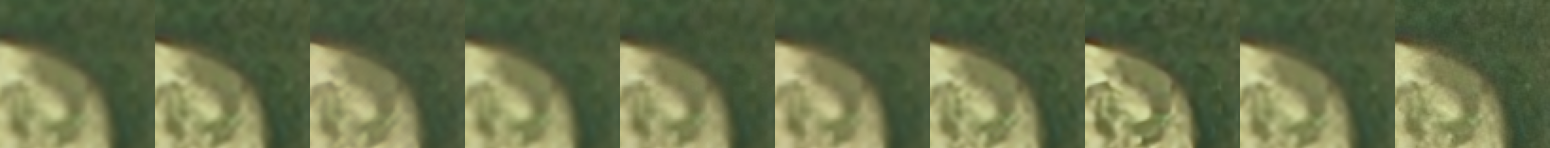}
\end{subfigure}
\begin{subfigure}[c]{\linewidth}
\includegraphics[width=\textwidth]{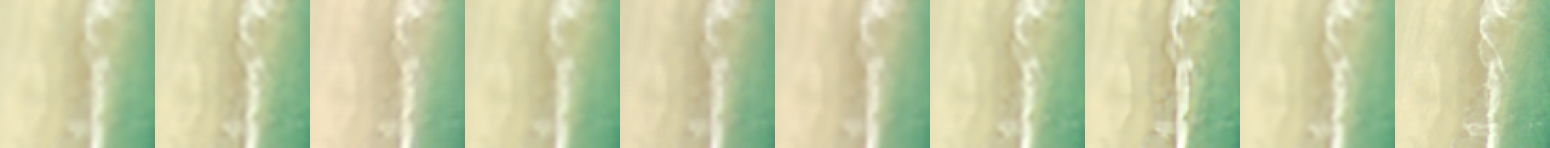}
\end{subfigure}
\begin{subfigure}[c]{\linewidth}
\includegraphics[width=\textwidth]{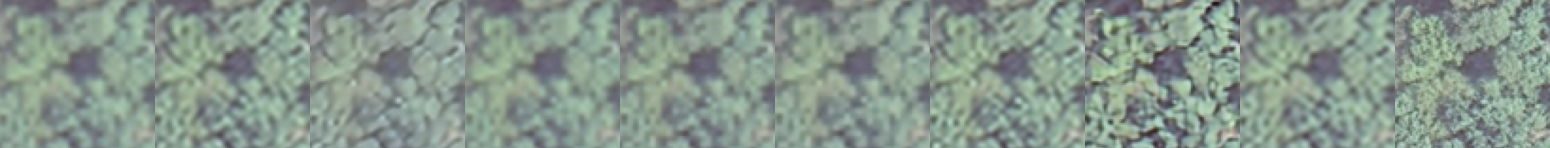}
\begin{subfigure}[c]{\linewidth}
\includegraphics[width=\textwidth]{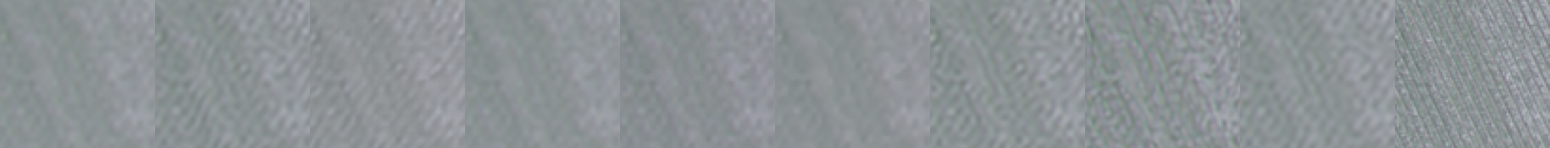}
\end{subfigure}
\end{subfigure}
\begin{subfigure}[c]{\linewidth}
\includegraphics[width=\textwidth]{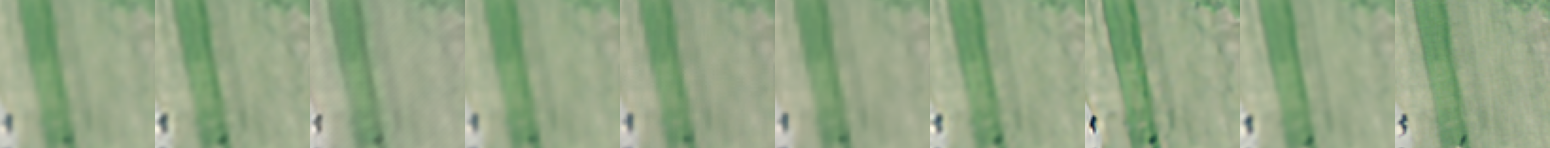}
\end{subfigure}
\begin{subfigure}[c]{\linewidth}
\includegraphics[width=\textwidth]{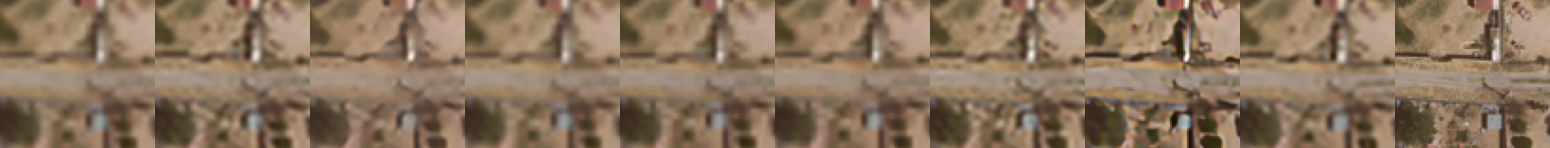}
\end{subfigure}
\begin{subfigure}[c]{\linewidth}
\includegraphics[width=\textwidth]{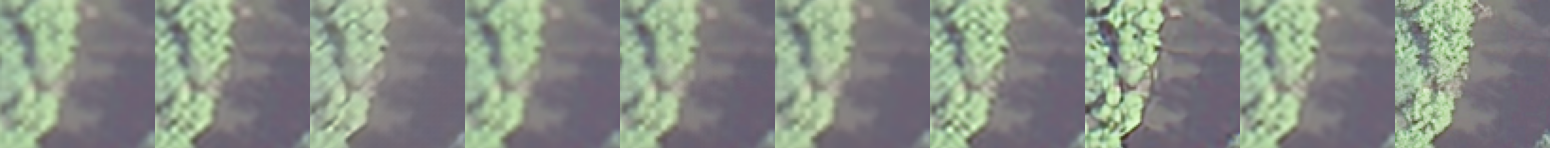}
\end{subfigure}
\begin{subfigure}[c]{\linewidth}
\includegraphics[width=\textwidth]{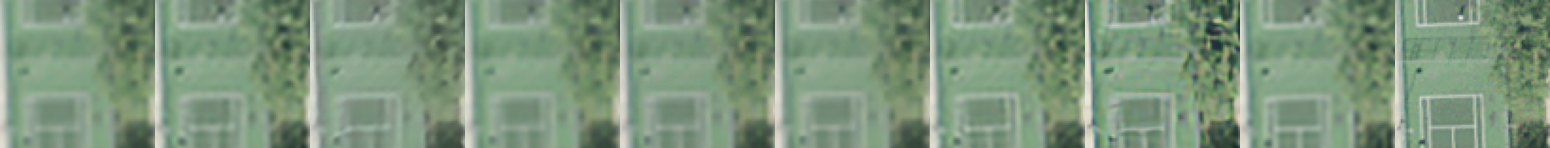}
\end{subfigure}
\begin{subfigure}[c]{\linewidth}
\includegraphics[width=\textwidth]{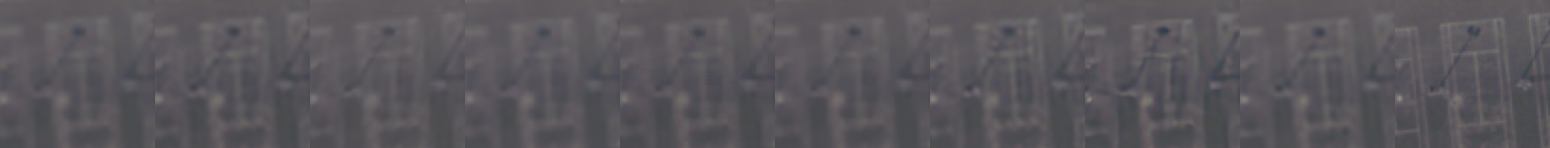}
\end{subfigure}
\begin{subfigure}[c]{\linewidth}
\includegraphics[width=\textwidth]{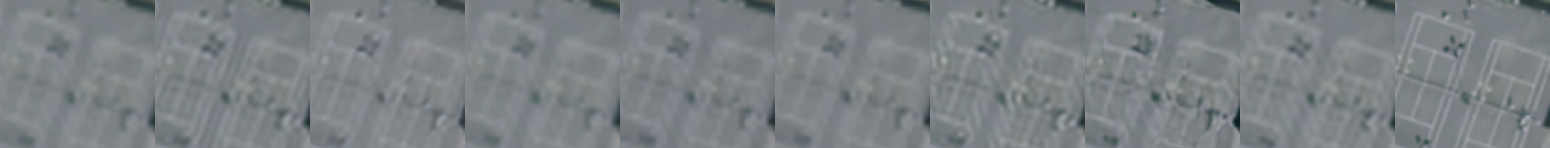}
\end{subfigure}
\caption{Examples (land and crops) of UCMerced images (crops of 256x256, with a zoom of x.40 i.e. 100x100) and each SR Algorithm output. \textit{LR} (corresponding to input on algorithms) is the downsampling of \textbf{HR} x3.}
\label{fig:srimages2}
\end{figure}

In our results for downsampling LR x3 (used for training and validation), we can qualitatively see in road, building and land crops shown in Figures \ref{fig:srimages1}-\ref{fig:srimages2} that FSRCNN, SRGAN and $MSRN_1$ present blurred output, similiar to the LR. For ESRGAN it presents a much sharper output however it seems to add extra noise in the edges. CAR however seems to acquire better results but it appears in some cases to be oversharpened (see tennis courts of Figure \ref{fig:srimages2} columns 8-10). In contrast, LIIF presents a better output with a slight blur.

\begin{figure}[H]
\centering
\begin{subfigure}[c]{\linewidth}
\begin{tabular}{lllllllll}
    \textit{Original} & \hspace{5pt} FSRCNN & \hspace{5pt} SRGAN & $MSRN_1$ & $MSRN_2$ & $MSRN_3$ &  ESRGAN & CAR & \hspace{18pt} LIIF 
\end{tabular}
\end{subfigure}
\begin{subfigure}[c]{\linewidth}
\includegraphics[width=\textwidth]{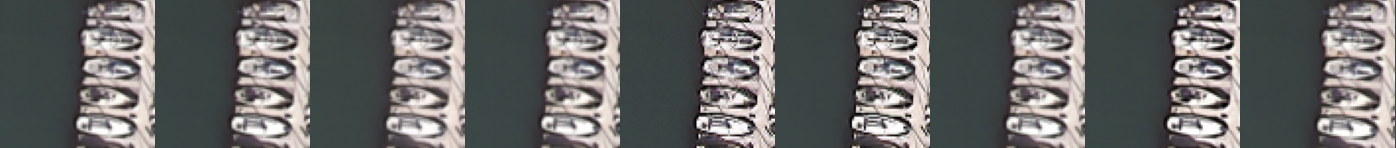}
\end{subfigure}
\begin{subfigure}[c]{\linewidth}
\includegraphics[width=\textwidth]{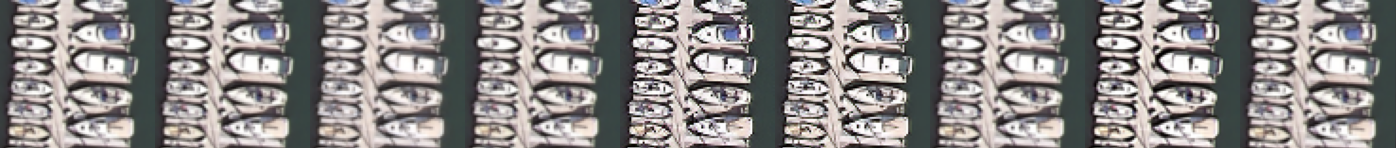}
\end{subfigure}
\begin{subfigure}[c]{\linewidth}
\includegraphics[width=\textwidth]{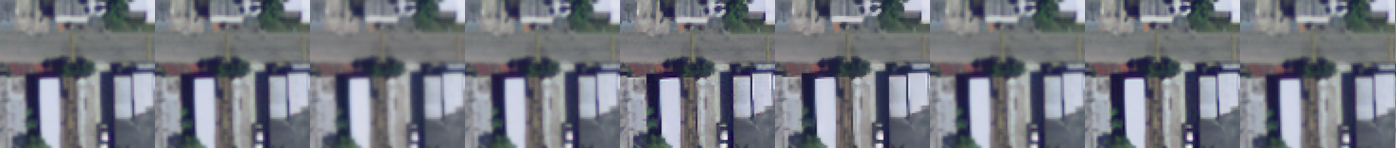}
\end{subfigure}
\begin{subfigure}[c]{\linewidth}
\includegraphics[width=\textwidth]{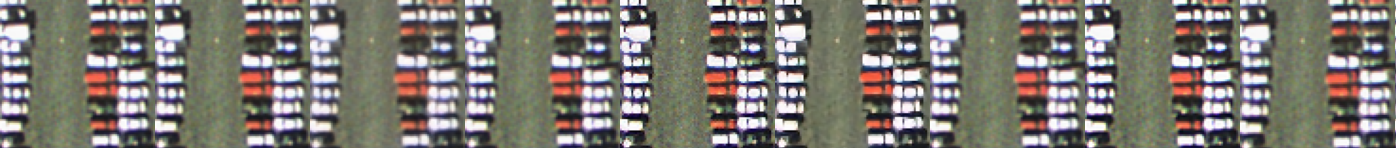}
\end{subfigure}
\begin{subfigure}[c]{\linewidth}
\includegraphics[width=\textwidth]{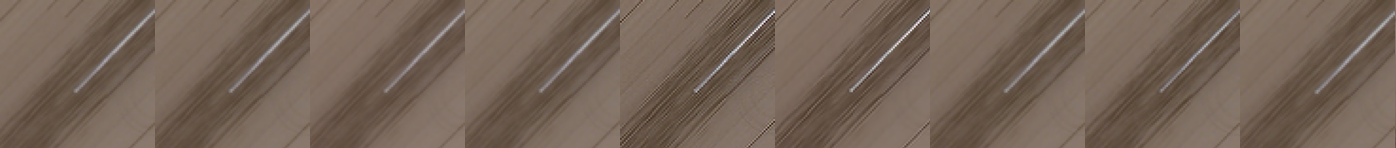}
\end{subfigure}
\begin{subfigure}[c]{\linewidth}
\includegraphics[width=\textwidth]{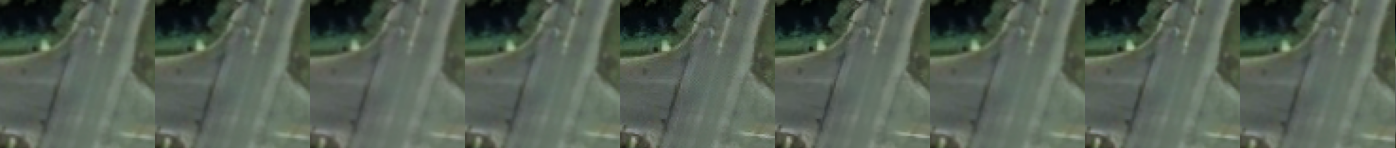}
\end{subfigure}
\begin{subfigure}[c]{\linewidth}
\includegraphics[width=\textwidth]{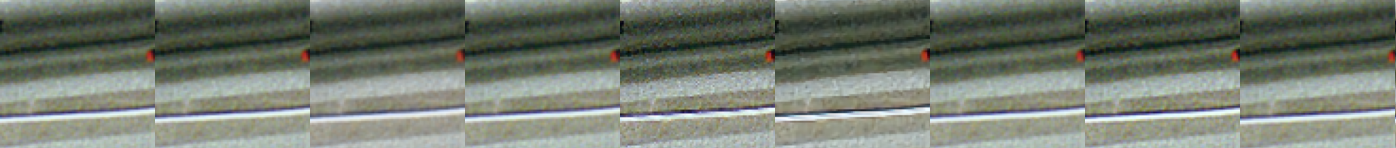}
\end{subfigure}
\begin{subfigure}[c]{\linewidth}
\includegraphics[width=\textwidth]{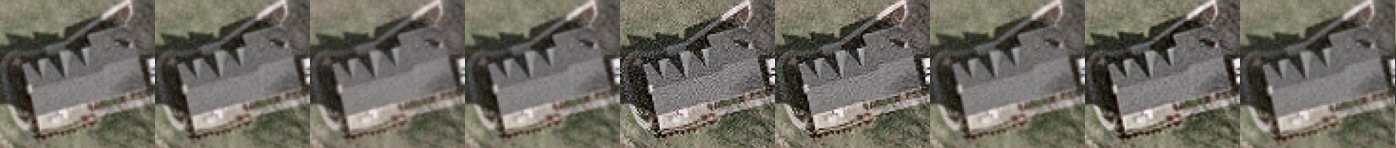}
\end{subfigure}
\begin{subfigure}[c]{\linewidth}
\includegraphics[width=\textwidth]{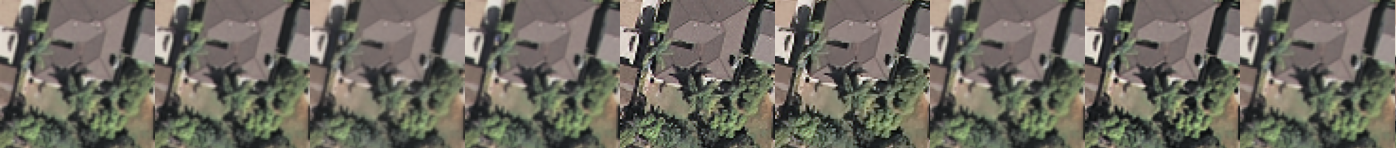}
\end{subfigure}
\begin{subfigure}[c]{\linewidth}
\includegraphics[width=\textwidth]{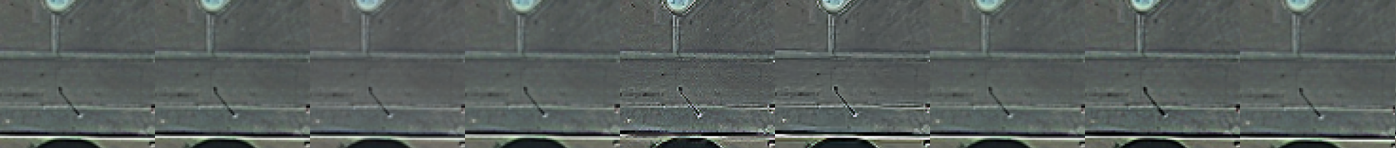}
\end{subfigure}
\begin{subfigure}[c]{\linewidth}
\includegraphics[width=\textwidth]{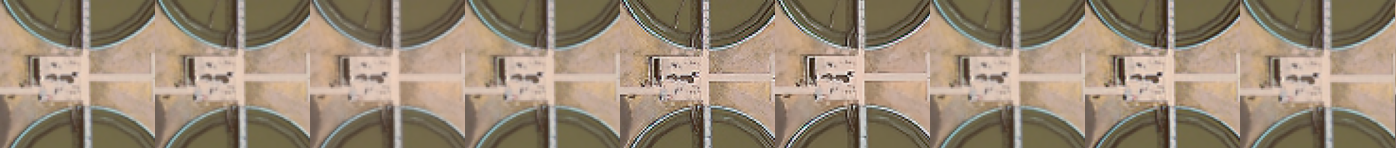}
\end{subfigure}
\caption{Examples (buildings and roads) of super-resolving original UCMerced images UCMerced images (crops of 256x256, with a zoom of x.40 i.e. 100x100) and each SR Algorithm output. Inputs (LR) without downsampling (corresponding to an upscaling of \textbf{HR} x3).}
\label{fig:srrimages1}
\end{figure}

\begin{figure}[H]
\centering
\begin{subfigure}[c]{\linewidth}
\begin{tabular}{lllllllll}
    \textit{Original} & \hspace{5pt} FSRCNN & \hspace{5pt} SRGAN & $MSRN_1$ & $MSRN_2$ & $MSRN_3$ &  ESRGAN & CAR & \hspace{18pt} LIIF 
\end{tabular}
\end{subfigure}
\begin{subfigure}[c]{\linewidth}
\includegraphics[width=\textwidth]{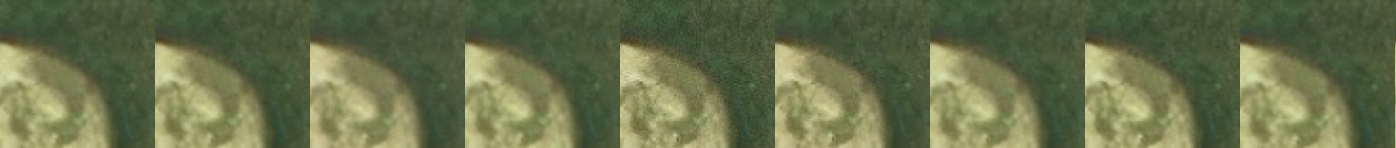}
\end{subfigure}
\begin{subfigure}[c]{\linewidth}
\includegraphics[width=\textwidth]{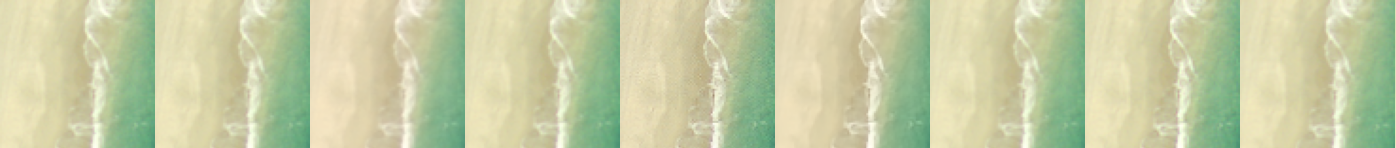}
\end{subfigure}
\begin{subfigure}[c]{\linewidth}
\includegraphics[width=\textwidth]{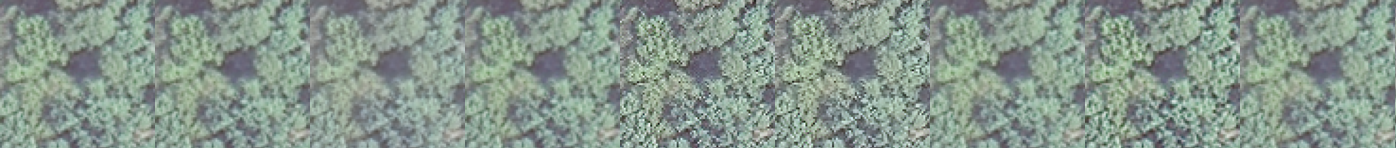}
\begin{subfigure}[c]{\linewidth}
\includegraphics[width=\textwidth]{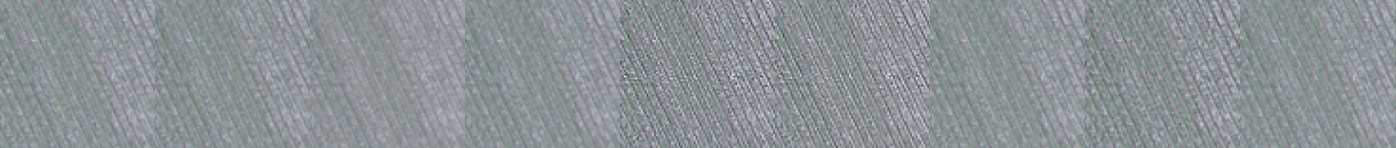}
\end{subfigure}
\end{subfigure}
\begin{subfigure}[c]{\linewidth}
\includegraphics[width=\textwidth]{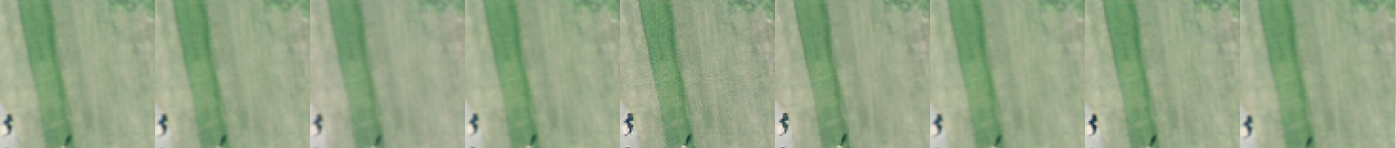}
\end{subfigure}
\begin{subfigure}[c]{\linewidth}
\includegraphics[width=\textwidth]{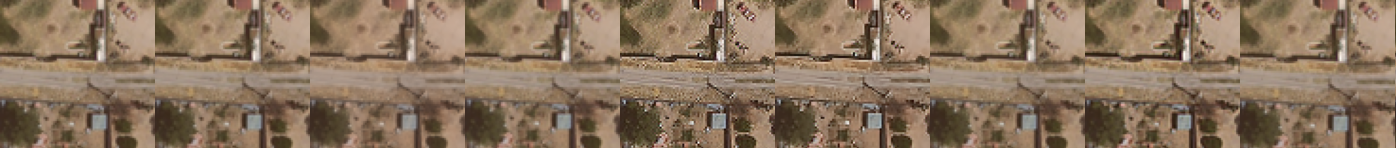}
\end{subfigure}
\begin{subfigure}[c]{\linewidth}
\includegraphics[width=\textwidth]{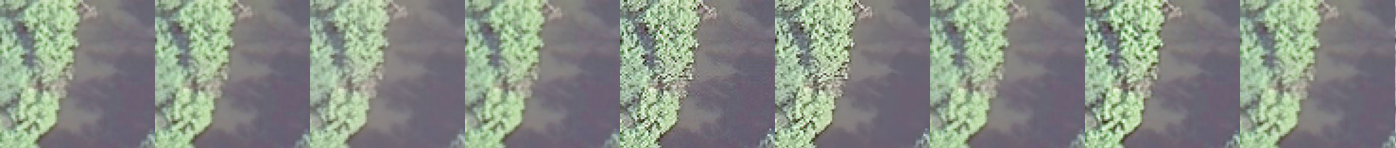}
\end{subfigure}
\begin{subfigure}[c]{\linewidth}
\includegraphics[width=\textwidth]{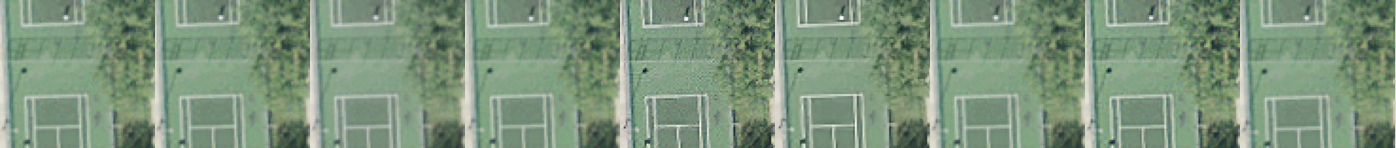}
\end{subfigure}
\begin{subfigure}[c]{\linewidth}
\includegraphics[width=\textwidth]{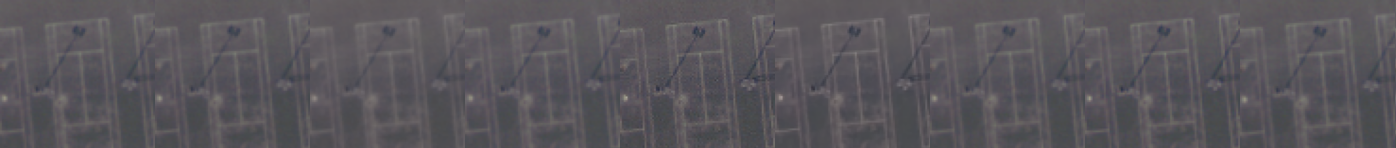}
\end{subfigure}
\begin{subfigure}[c]{\linewidth}
\includegraphics[width=\textwidth]{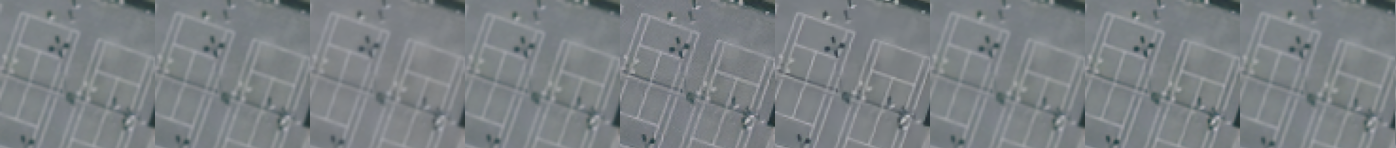}
\end{subfigure}
\caption{Examples (land and crops) of super-resolving original UCMerced images (crops of 256x256, with a zoom of x.40 i.e. 100x100) and each SR Algorithm output. Inputs (LR) without downsampling (corresponding to an upscaling of \textbf{HR} x3).}
\label{fig:srrimages2}
\end{figure}

In Figures \ref{fig:srrimages1}-\ref{fig:srrimages2} we super-resolve the original UCMerced images x3 and we can observe some alrogithms such as FSRCNN, SRGAN, $MSRN_1$, ESRGAN and LIIF present a similar (blurred) output to the GT, while others such as $MSRN_2$, $MSRN_3$ and CAR present a higher noise and oversharpening of borders, trying to enhance features of the image (here attempting to generate features with a GSD lower than 30 cm/px). 

\begin{table}[h!]
    \centering
    \scriptsize
    \end{table}







\begin{table}[h!]
    \centering
    \scriptsize	
    \hspace*{-1.75in}
\begin{minipage}{.56\linewidth}
\rotatebox[origin=c]{90}{x2}
\csvautotabular[respect all]{similarity_x2_blurfalse.csv}
\end{minipage}
\begin{minipage}{.44\linewidth}
\hspace{0.5in}
\rotatebox[origin=c]{90}{x2 + blur}
\csvautotabular[respect all]{similarity_x2_blurtrue.csv}
\end{minipage}\\
    \hspace*{-1.75in}
\begin{minipage}{.56\linewidth}
\rotatebox[origin=c]{90}{x3}
\csvautotabular[respect all]{similarity_x3_blurfalse.csv}
\end{minipage}
\begin{minipage}{.44\linewidth}
\hspace{0.5in}
\rotatebox[origin=c]{90}{x3 + blur}
\csvautotabular[respect all]{similarity_x3_blurtrue.csv}
\end{minipage}\\
    \hspace*{-1.75in}
\begin{minipage}{.56\linewidth}
\rotatebox[origin=c]{90}{x4}
\csvautotabular[respect all]{similarity_x3_blurtrue.csv}
\end{minipage}
\begin{minipage}{.44\linewidth}
\hspace{0.5in}
\rotatebox[origin=c]{90}{x4+blur}
\csvautotabular[respect all]{similarity_x4_blurtrue.csv}
\end{minipage} 
\caption{Mean Full-Reference Metrics on Super-Resolution of downsampled inputs in UCMerced-380.}
    \label{tab:sisr_fullreference}
\end{table}







\begin{table}[H]
    \centering
    \scriptsize	
    \hspace*{-1.25in}
\begin{minipage}{.56\linewidth}
    \rotatebox[origin=c]{90}{x2}
    \csvautotabular[respect all]{noise_x2_blurfalse.csv}
    \end{minipage}
    \hspace*{.25in}
\begin{minipage}{.44\linewidth}
    \rotatebox[origin=c]{90}{x2+blur}
    \csvautotabular[respect all]{noise_x2_blurtrue.csv}
    \end{minipage} \\ 
    \hspace*{-1.25in}
\begin{minipage}{.56\linewidth}
    \rotatebox[origin=c]{90}{x3}
    \csvautotabular[respect all]{noise_x3_blurfalse.csv}
    \end{minipage} 
    \hspace*{.25in}
\begin{minipage}{.44\linewidth}
    \rotatebox[origin=c]{90}{x3+blur}
    \csvautotabular[respect all]{noise_x3_blurtrue.csv}
    \end{minipage} \\
    \hspace*{-1.25in}
\begin{minipage}{.56\linewidth}
    \rotatebox[origin=c]{90}{x4}
    \csvautotabular[respect all]{noise_x4_blurfalse.csv}
    \end{minipage} 
    \hspace*{.25in}
    \begin{minipage}{.44\linewidth}
    \rotatebox[origin=c]{90}{x4+blur}
    \csvautotabular[respect all]{noise_x4_blurtrue.csv}
    \end{minipage} 
    \caption{Mean No-Reference Noise (SNR) and Sharpness (RER, MTF, FWHM) Metrics on Super-Resolution of downsampled inputs in UCMerced-380.}
    \label{tab:sisr_snr}
\end{table}

\begin{figure}[H]
\centering
\begin{subfigure}[c]{0.48\textwidth}
\includegraphics[width=\textwidth]{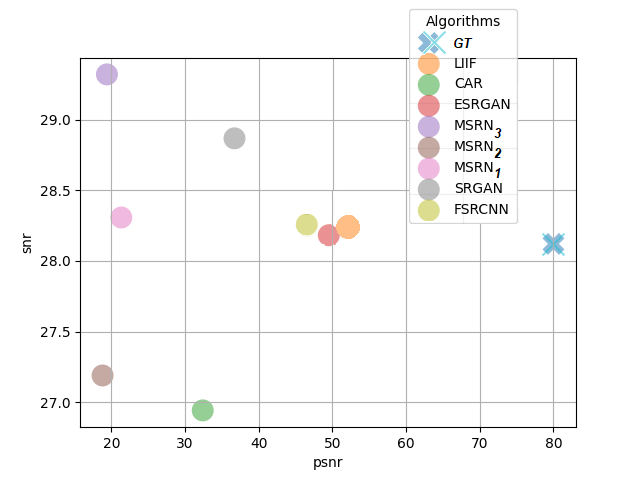}
\end{subfigure}
\begin{subfigure}[c]{0.48\textwidth}
\includegraphics[width=\textwidth]{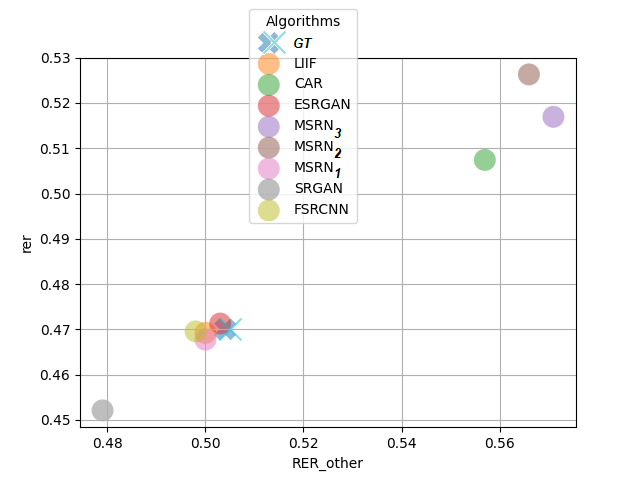}
\end{subfigure}
\begin{subfigure}[c]{0.48\textwidth}
\includegraphics[width=\textwidth]{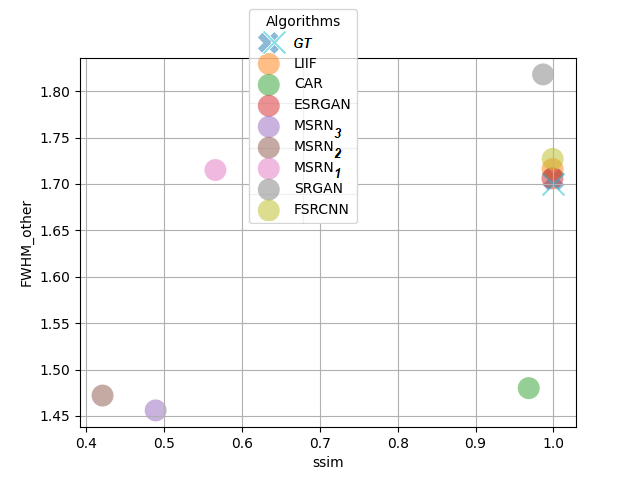}
\end{subfigure}
\begin{subfigure}[c]{0.48\textwidth}
\includegraphics[width=\textwidth]{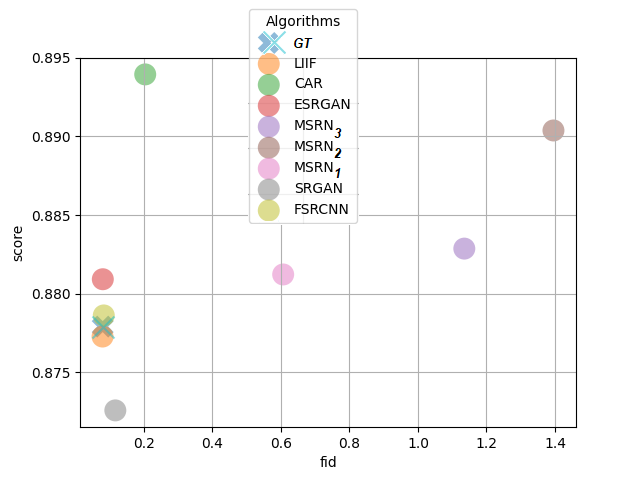}
\end{subfigure}
\caption{Scatter plots of metric comparison on Super Resolving (x4) UCMerced dataset.}
\label{fig:scatters}
\end{figure}

Above (Figure \ref{fig:scatters}) we demonstrate validity of some of our metric results by comparing them with each homologous measurement, namely, the ones measuring similar or same properties. Here we compared QMRNet's $snr\downarrow$ and PSNR$\uparrow$. These measure the quantity of noise over information of the image. The first subplot shows an anticorrelation ($\swarrow$) on algorithm values in these 2 metrics, with LIIF being closest to the HR (GT) and CAR, $MSRN_2$ and $MSRN_3$ having both lowest $snr$ (best) and PSNR (worst). For the case of QMRNet's $rer\uparrow$ and measured $RER_{other}\uparrow$ (which corresponds to the RER that measure diagonal contours), there is a positive correlation ($\nearrow$), with CAR, $MSRN_2$ and $MSRN_3$ outperforming the rest of algorithms. We also compared $FWHM_{other}\downarrow$ and SSIM$\uparrow$ to see how well each algorithm performs when evaluating diagonal contour width as well as structural similarity, and it appears that $MSRN_2$, $MSRN_3$, and CAR get lowest (best) $FWHM$ and most algorithms getting same values of SSIM as original GT images (unchanged). In the last subplot we compared the QMRNet's $score\uparrow$ (composed by weighted mean of QMRNet's $blur$, $rer$, $snr$, $GSD$ and $F$) and FID$\downarrow$, which measures Frechet distribution distance between images. Here $MSRN_2$, $MSRN_3$ and CAR show highest $score$ with higher (worse) FID, while most algorithms are close to the original image (almost unchanged). Note that in these plots we super-resolve x4 the original image so that Full-Reference metrics can only compare with the original image (thus, there is no downsampling of inputs so that the HR would be equivalent to the LR input), here we need to consider how algorithms actually perform in metrics that can evaluate better than the original image of 30 cm/px.


\subsection{Results on QMRloss: Optimizing Image Super-Resolution}


In this section, we integrated the aforementioned QMRLoss as an ad-hoc strategy for optimizing SR algorithms\footnote{Check QMRLoss optimization use case in \url{https://github.com/dberga/iquaflow-qmr-loss}}. For this case, we integrated different loss methods (L1, L2 and BCE) as QMRLoss in different modifiers in MSRN training. MSRN architecture includes a noise addition when training upon a specific LR, here we added the QMRLoss to the total loss calculation, namely, summed to the autoencoder Spatial (adversarial) Loss and VGG (perceptual) Loss. This QMRLoss mechanism will allow MSRN (and any other algorithm integrated with) to avoid any quality mismatch considering several metrics that measure distortions simultaneously. 
\begin{figure}[H]
\centering
\vspace{-24pt}
\begin{subfigure}[c]{0.48\textwidth}
\caption*{PSNR}
\includegraphics[width=\textwidth]{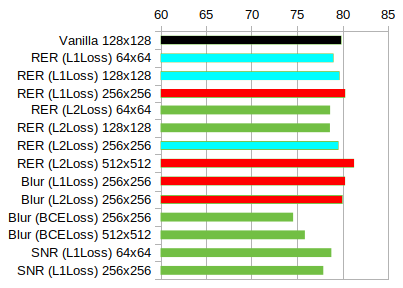}
\end{subfigure}
\begin{subfigure}[c]{0.48\textwidth}
\caption*{SSIM}
\includegraphics[width=\textwidth]{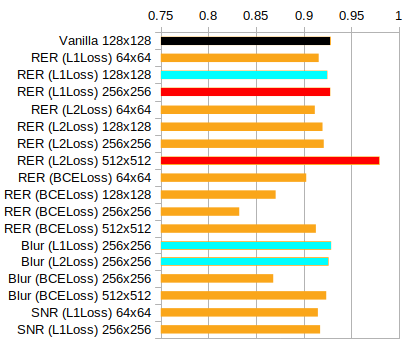}
\end{subfigure}
\begin{subfigure}[c]{0.48\textwidth}
\caption*{FID}
\includegraphics[width=\textwidth]{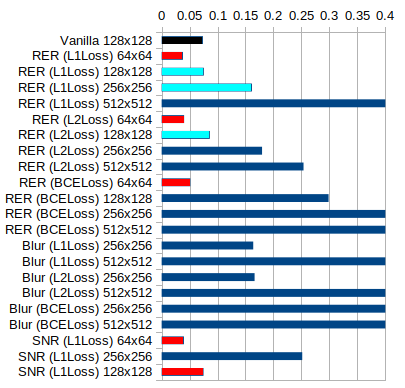}
\end{subfigure}
\begin{subfigure}[c]{0.48\textwidth}
 Mean \textcolor{green}{PSNR}, \textcolor{orange}{SSIM} and \textcolor{blue}{FID} are selected from last 10 epochs. Outperforming values (from vanilla MSRN) are marked in \textcolor{red}{red}, and top-3 around/below vanilla are marked in \textcolor{cyan}{cyan}. 
\end{subfigure}
\caption{Validation of QMRLoss optimizing MSRN in Super Resolution of Inria-AILD-180. Note that the training / validation regime has been done over inria-AILD-180 with 80\%20\% splits and crops set to 64x64, 128x128, 256x256, 512x512.}
\label{fig:trainmsrn1}
\end{figure}

In Figure \ref{fig:trainmsrn1} we show that several strategies such as QMRLoss using $rer$ (and L1Loss) gets better results than vanilla MSRN in PSNR, SSIM and FID metrics. Here PSNR improves with QMRNet using L1 loss and crops of 256x256 as well as with L2 loss with 512x512. It also improves with $blur$ metric both with L1 and L2 Loss on 256x256 crops. SSIM improves with L1 Loss in QMRNet that uses RER and significatively (almost 1.0) with $rer$ L2 loss with crops of 512x512. For FID, using QMRNet improves MSRN with $rer$ and all types of losses (L1, L2 and BCE) using crops of 64x64, here as well using QMRNet with $snr$ metric and L1 Loss, using crops of 64x64 and 128x128.

We also tested our MSRN+QMRLoss' generated images with most of our Full-Reference and No-Reference metrics in UCMerced-380 dataset (outside inria-AILD's training and validation distribution) with crops of 256x256. Here vanilla's MSRN gets worse results for $sigma\downarrow$, $rer\uparrow$, SSIM$\uparrow$, SWD $\downarrow$, FID $\downarrow$, snr$_{Mdn}$ $\downarrow$, RER mean of X\&Y $\uparrow$, MTF mean of X\&Y $\uparrow$ and FWHM mean of X\&Y $\downarrow$ in comparison with the optimized QMR*snr256, QMR*rer256 and QMR*blur256. Here QMRLoss$_{L1}$ has been able to adapt better when generating contours and predicting blurred objects on testing distinct shapes from the original training.


\begin{table}[H]
    \centering
    \scriptsize
    \hspace*{-1.0in}
    \begin{minipage}{\linewidth}
    \rotatebox[origin=c]{90}{x3}
    \csvautotabular[respect all]{qmrloss_x3_blurfalse.csv}
    \end{minipage}\\
    \caption{Test metrics on Super-Resolution (downsampling inputs x3) using QMRNets (MSRN backbone) in UCMerced-380 (note that here we are adapting QMRNet with distinct input crops from inria-AILD, while this test dataset are around 256x256). *$QMRloss_{L1}$ computation over inria-AILD-train on distinct QMRNets (for blur, rer and snr) using crops ($R$) of 256x256.}
    \label{tab:sisr_qmrloss}
\end{table}

\section*{\uppercase{Conclusions}}

In this study, we implement an open-source tool (integrated in the IQUAFLOW framework) developed for assessing quality and modifying EO images. We propose a network architecture (QMRNet using VGG19) that predicts the amount of distortion for each parameter as a no-reference metric. We also benchmark distinct super-resolution algorithms and datasets with both full-reference and no-reference metrics, and propose a novel mechanism for optimizing super-resolution training regimes using QMRLoss, integrating QMRNet metrics with SR algorithm objectives.

On assessing image quality of datasets we observe similar overall score for most datasets, with dissimilarities in scores of snr and rer. On assessing single image super resolution we see significatively better results for CAR, LIIF, $MSRN_2$ and $MSRN_3$. On optimizing MSRN with QMRLoss (snr, rer and blur) improves results on both full-reference and no-reference metrics with respect default vanilla MSRN.

We have to pinpoint that our proposed method can be applied to any other kind of distortion or modification. QMRNet allows to predict any parameter of the image and also several parameters simultaneously. For instance, training QMRNet to assess compression parameters could be another use case of interest. We also tested the usage of QMRNet as Loss for optimizing SR of MSRN, but it could be extended with distinct algorithm architectures and uses, as QMRLoss allows to reverse or denoise any modification on the original image. In addition, it is also possible to implement a variation to the QMRLoss objective by forcing the head to be on a specific interval (with maximum quality and minimal distortion for each parameter) so that the algorithm maximizes toward a specific metric or score regardless of the output of QMRNet on GT.

\section*{\uppercase{Acknowledgements}}

The project was financed by the Ministry of Science and Innovation (MICINN) and by the European Union within the framework of FEDER RETOS-Collaboration of the State Program of Research (RTC2019-007434-7), Development and Innovation Oriented to the Challenges of Society, within the State Research Plan Scientific and Technical and Innovation 2017-2020, with the main objective of promoting technological development, innovation and quality research.

\bibliographystyle{apalike}
{\small
\bibliography{remotesensing}}

\begin{figure}[h!]
\centering
\begin{subfigure}[c]{\linewidth}
\begin{tabular}{llllllll}
    \textit{Original} & \hspace{15pt} vanilla &  $QMR_{\sigma,256}$ & $QMR_{\sigma,512*}$ & $QMR_{rer,256}$ & $QMR_{rer,512*}$ &  $QMR_{snr,256}$ & $QMR_{snr,512}$
\end{tabular}
\normalsize
\end{subfigure}
\begin{subfigure}[c]{\linewidth}
\includegraphics[width=\textwidth]{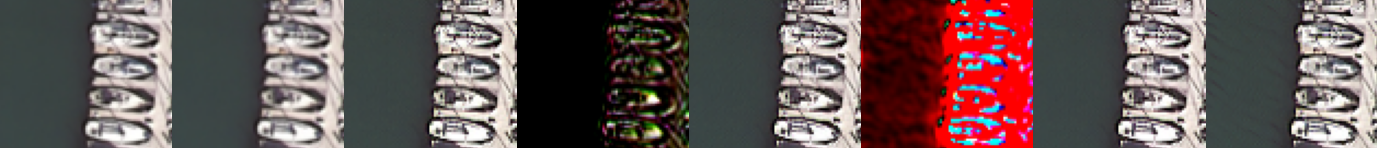}
\end{subfigure}
\begin{subfigure}[c]{\linewidth}
\includegraphics[width=\textwidth]{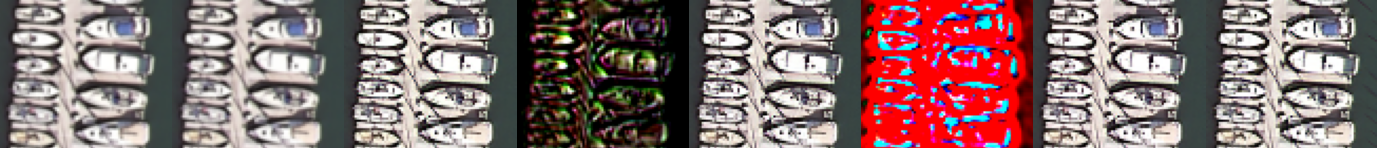}
\end{subfigure}
\begin{subfigure}[c]{\linewidth}
\includegraphics[width=\textwidth]{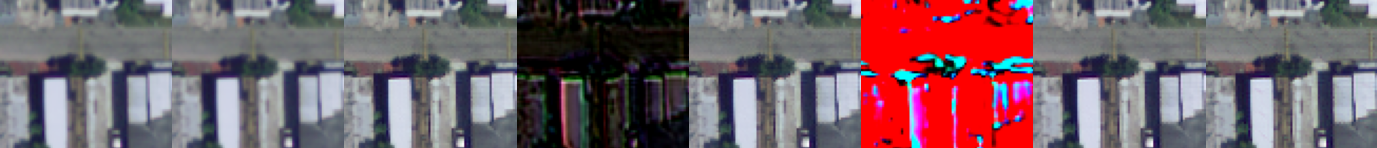}
\end{subfigure}
\begin{subfigure}[c]{\linewidth}
\includegraphics[width=\textwidth]{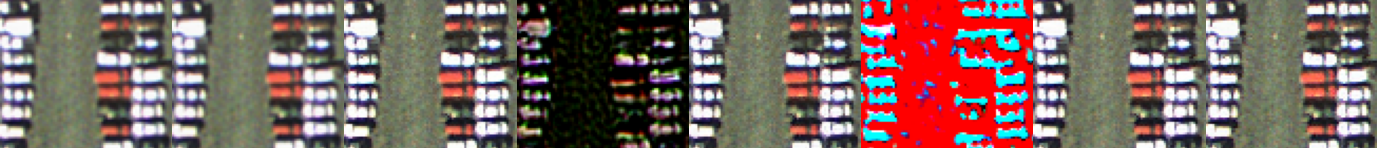}
\end{subfigure}
\begin{subfigure}[c]{\linewidth}
\includegraphics[width=\textwidth]{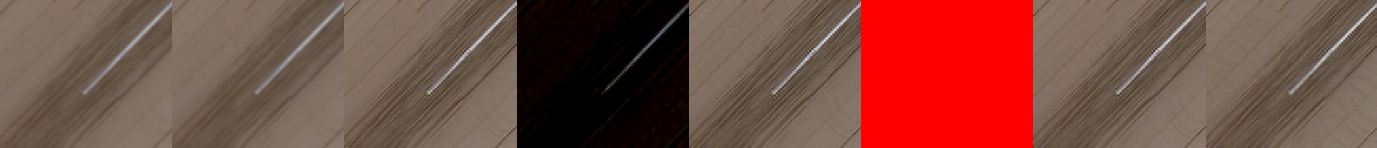}
\end{subfigure}
\begin{subfigure}[c]{\linewidth}
\includegraphics[width=\textwidth]{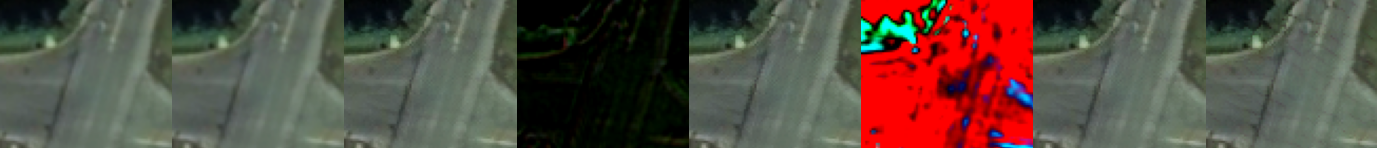}
\end{subfigure}
\begin{subfigure}[c]{\linewidth}
\includegraphics[width=\textwidth]{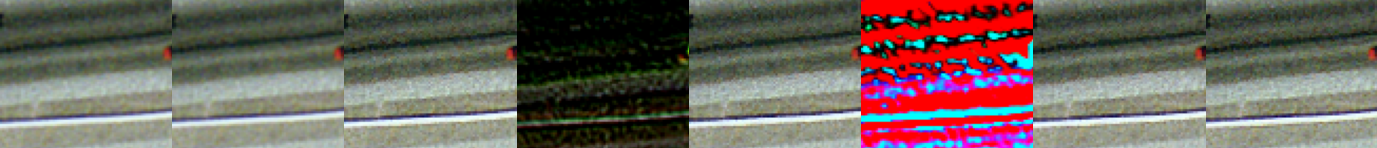}
\end{subfigure}
\begin{subfigure}[c]{\linewidth}
\includegraphics[width=\textwidth]{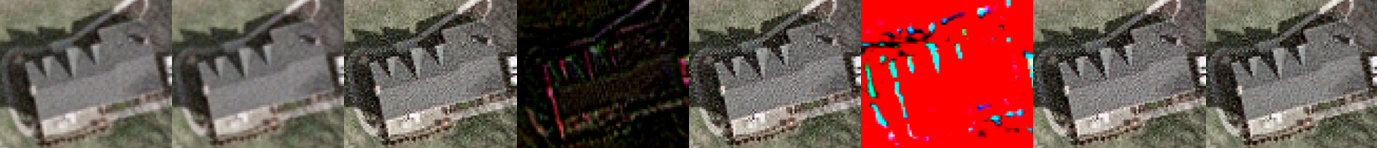}
\end{subfigure}
\begin{subfigure}[c]{\linewidth}
\includegraphics[width=\textwidth]{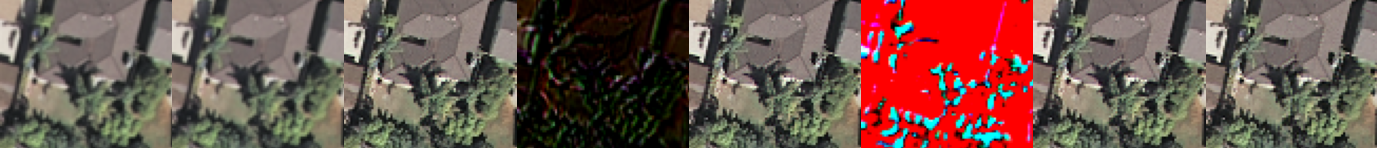}
\end{subfigure}
\begin{subfigure}[c]{\linewidth}
\includegraphics[width=\textwidth]{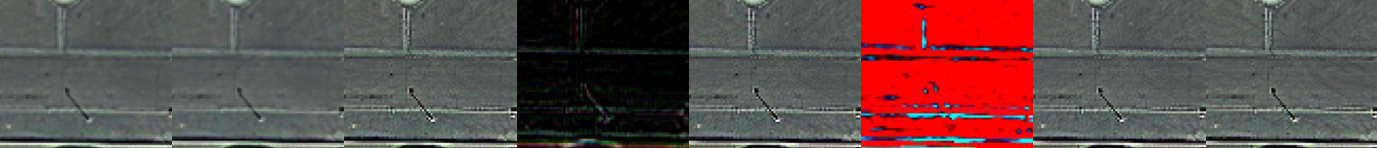}
\end{subfigure}
\begin{subfigure}[c]{\linewidth}
\includegraphics[width=\textwidth]{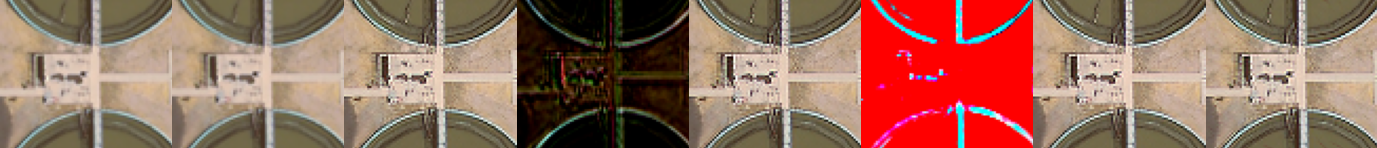}
\end{subfigure}
\caption{Examples (buildings and roads) of UCMerced images (crops of 256x256, with a zoom of x.40 i.e. 100x100) and each QMRNet Algorithm output (QMRLoss$_{L1}$). \textit{LR} (corresponding to input on algorithms) is the downsampling of \textbf{HR} x3. *For $QMR_{512}$ images input is upscaled to 512x512 using a circular padding, thus, high-level features (from these QMRNets) activity are added in these columns.}
\label{fig:qmrimages1post}
\end{figure}

\begin{figure}[H]
\centering
\begin{subfigure}[c]{\linewidth}
\begin{tabular}{llllllll}
    \textit{Original} & \hspace{15pt} vanilla & $QMR_{\sigma,256}$ & $QMR_{\sigma,512*}$ & $QMR_{rer,256}$ & $QMR_{rer,512*}$ &  $QMR_{snr,256}$ & $QMR_{snr,512}$
\end{tabular}
\normalsize
\end{subfigure}
\begin{subfigure}[c]{\linewidth}
\includegraphics[width=\textwidth]{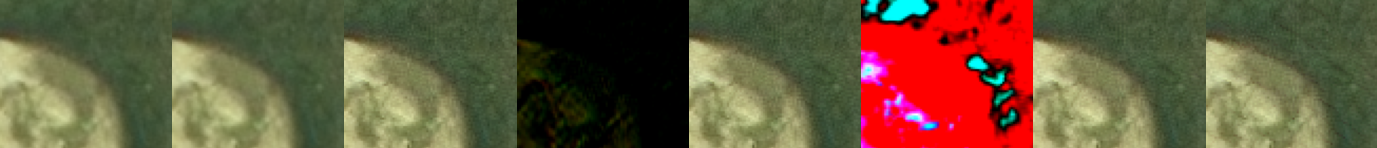}
\end{subfigure}
\begin{subfigure}[c]{\linewidth}
\includegraphics[width=\textwidth]{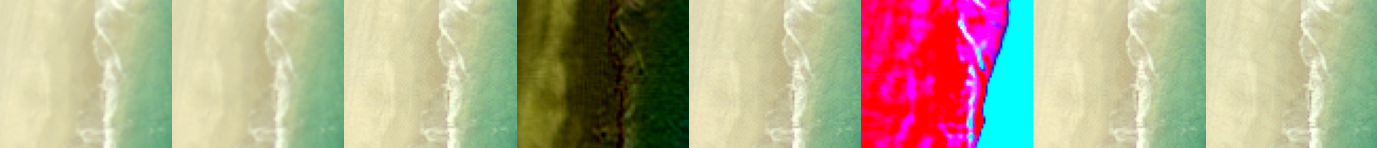}
\end{subfigure}
\begin{subfigure}[c]{\linewidth}
\includegraphics[width=\textwidth]{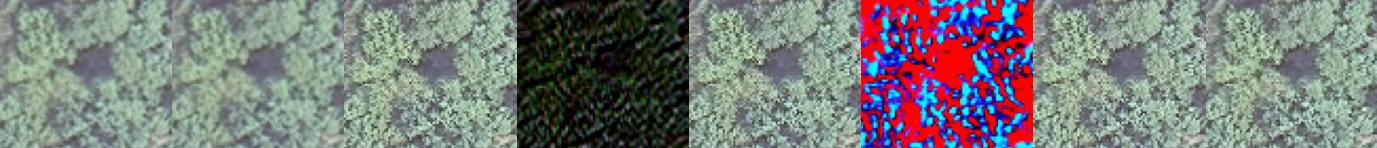}
\begin{subfigure}[c]{\linewidth}
\includegraphics[width=\textwidth]{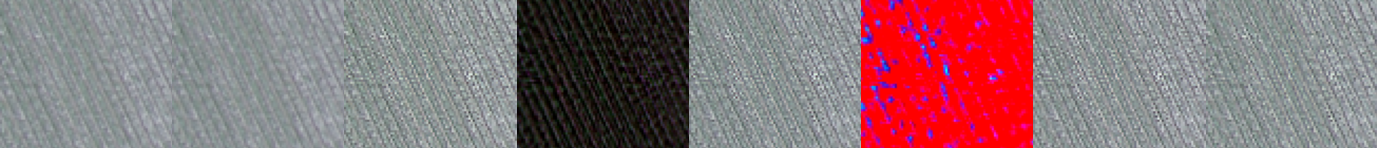}
\end{subfigure}
\end{subfigure}
\begin{subfigure}[c]{\linewidth}
\includegraphics[width=\textwidth]{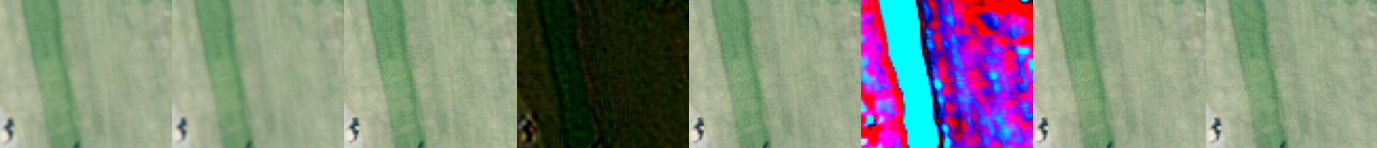}
\end{subfigure}
\begin{subfigure}[c]{\linewidth}
\includegraphics[width=\textwidth]{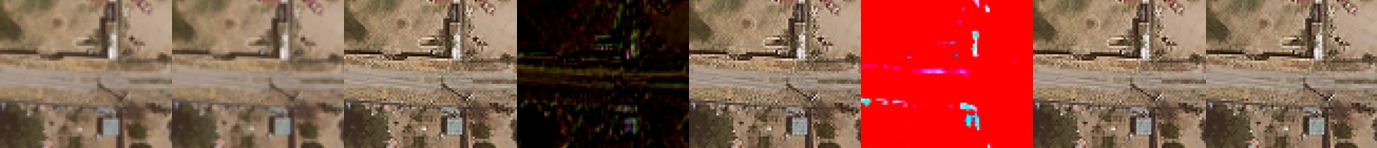}
\end{subfigure}
\begin{subfigure}[c]{\linewidth}
\includegraphics[width=\textwidth]{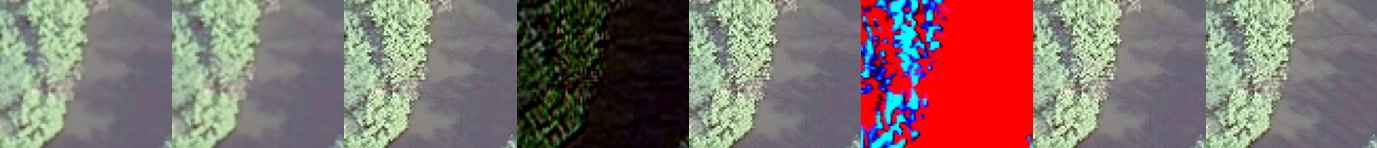}
\end{subfigure}
\begin{subfigure}[c]{\linewidth}
\includegraphics[width=\textwidth]{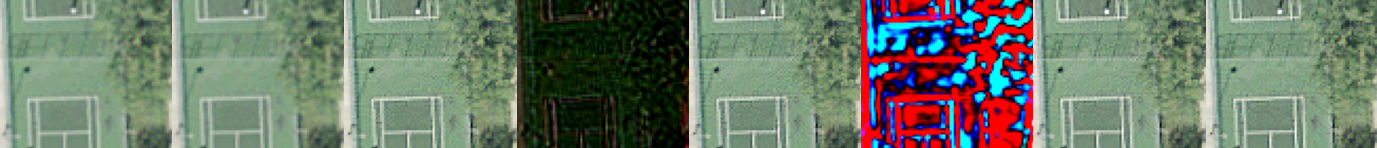}
\end{subfigure}
\begin{subfigure}[c]{\linewidth}
\includegraphics[width=\textwidth]{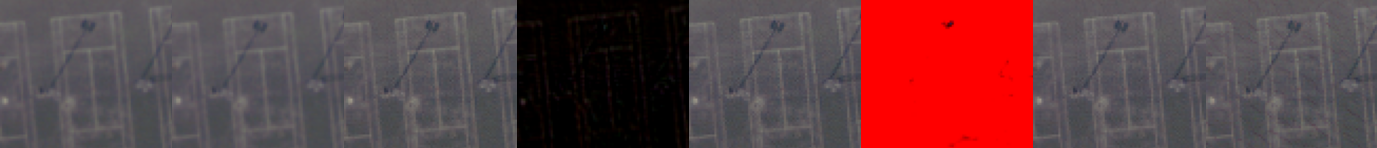}
\end{subfigure}
\begin{subfigure}[c]{\linewidth}
\includegraphics[width=\textwidth]{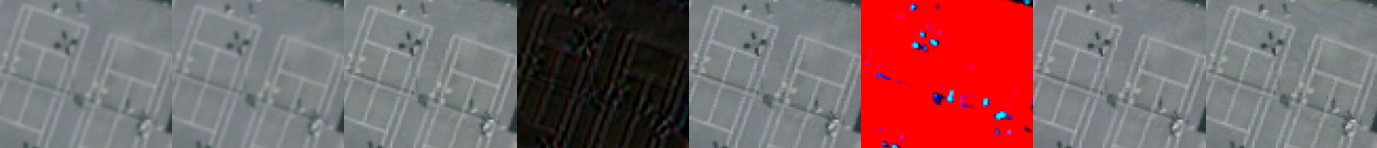}
\end{subfigure}
\caption{Examples (land and crops) of UCMerced images UCMerced images (crops of 256x256, with a zoom of x.40 i.e. 100x100) and each QMRNet Algorithm output (QMRLoss$_{L1}$). \textit{LR} (corresponding to input on algorithms) is the downsampling of \textbf{HR} x3. *For $QMR_{512}$ images input is upscaled to 512x512 using a circular padding, thus, high-level features (from these QMRNets) activity are added in these columns.}
\label{fig:qmrimages2post}
\end{figure}

\begin{figure}[H]
\centering
\begin{subfigure}[c]{\linewidth}
\small
\begin{tabular}{lllllllll}
    \textit{LR} & \hspace{18pt} vanilla & \hspace{8pt} $QMR_{\sigma,256}$ & \hspace{2pt} $QMR_{\sigma,512*}$ & \hspace{-6pt} $QMR_{rer,256}$  & \hspace{-6pt} $QMR_{rer,512*}$ &  \hspace{-6pt} $QMR_{snr,256}$ & \hspace{-6pt} $QMR_{snr,512*}$ & \hspace{8pt} \textbf{HR} 
\end{tabular}
\normalsize
\end{subfigure}
\begin{subfigure}[c]{\linewidth}
\includegraphics[width=\textwidth]{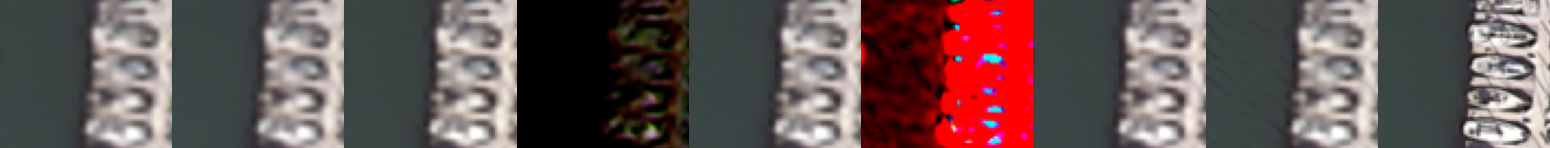}
\end{subfigure}
\begin{subfigure}[c]{\linewidth}
\includegraphics[width=\textwidth]{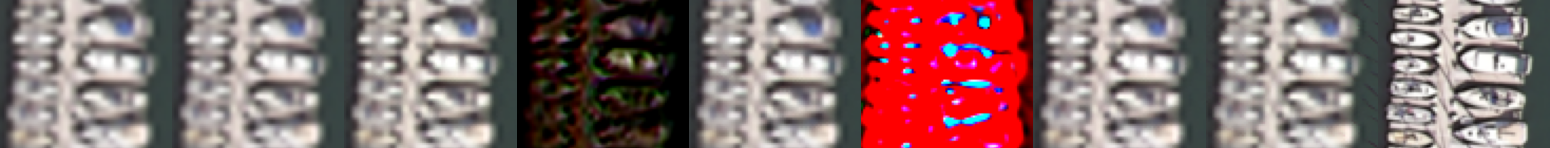}
\end{subfigure}
\begin{subfigure}[c]{\linewidth}
\includegraphics[width=\textwidth]{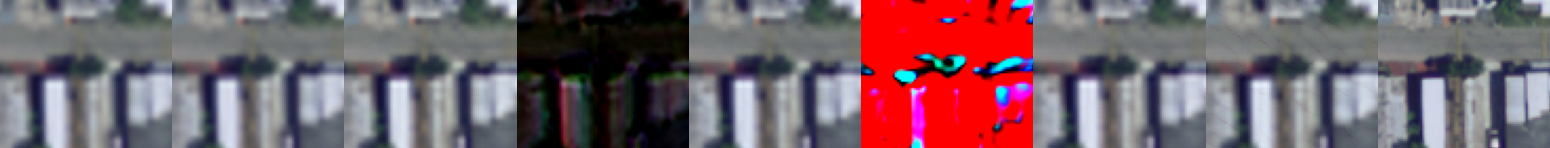}
\end{subfigure}
\begin{subfigure}[c]{\linewidth}
\includegraphics[width=\textwidth]{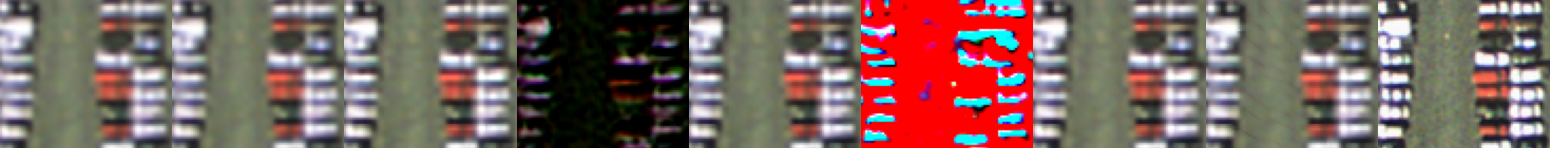}
\end{subfigure}
\begin{subfigure}[c]{\linewidth}
\includegraphics[width=\textwidth]{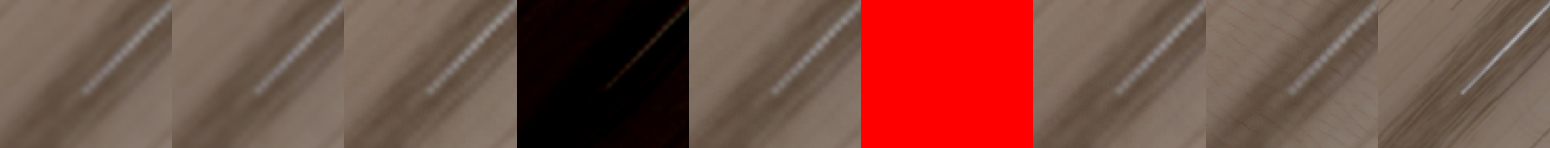}
\end{subfigure}
\begin{subfigure}[c]{\linewidth}
\includegraphics[width=\textwidth]{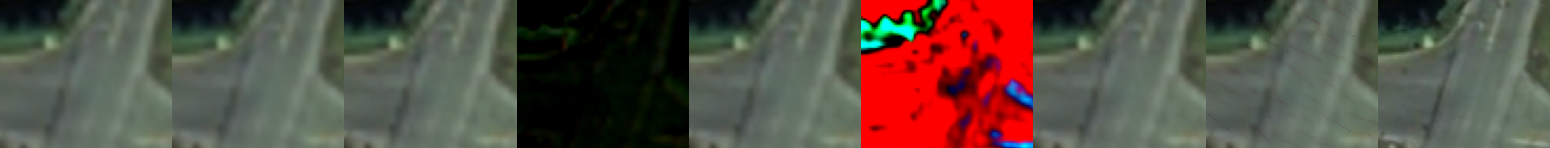}
\end{subfigure}
\begin{subfigure}[c]{\linewidth}
\includegraphics[width=\textwidth]{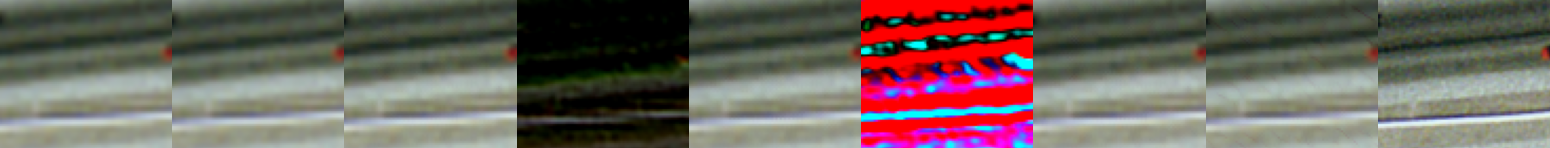}
\end{subfigure}
\begin{subfigure}[c]{\linewidth}
\includegraphics[width=\textwidth]{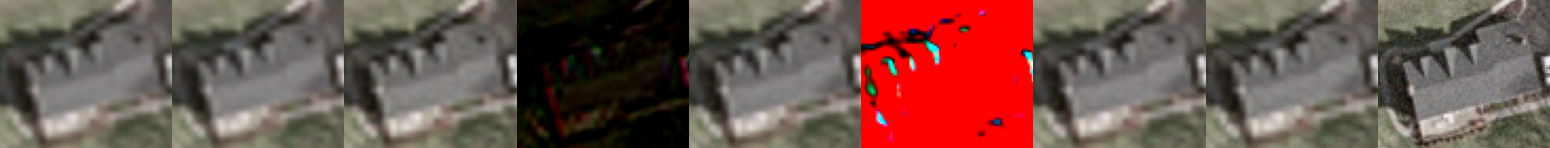}
\end{subfigure}
\begin{subfigure}[c]{\linewidth}
\includegraphics[width=\textwidth]{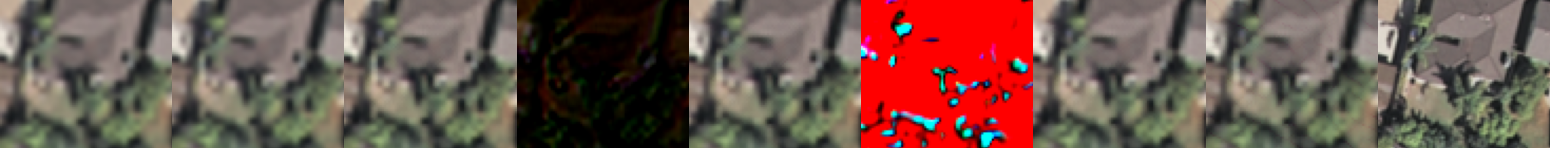}
\end{subfigure}
\begin{subfigure}[c]{\linewidth}
\includegraphics[width=\textwidth]{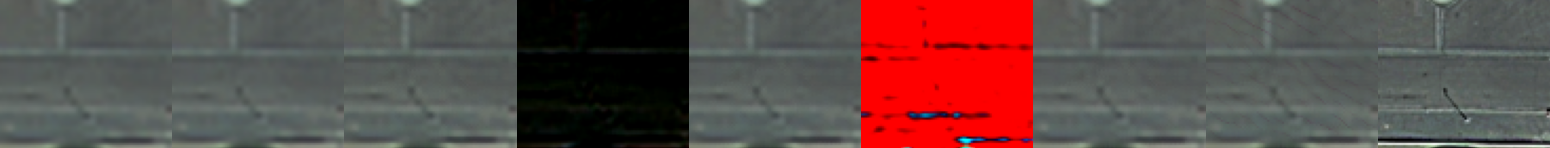}
\end{subfigure}
\begin{subfigure}[c]{\linewidth}
\includegraphics[width=\textwidth]{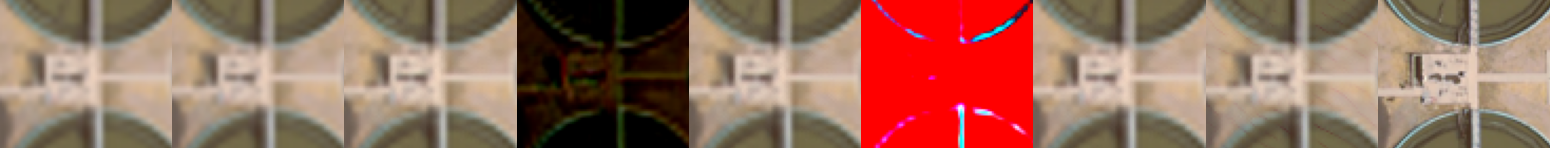}
\end{subfigure}
\caption{Examples (buildings and roads) of UCMerced images (crops of 256x256, with a zoom of x.40 i.e. 100x100) and each QMRNet Algorithm output (QMRLoss$_{L1}$). \textit{LR} (corresponding to input on algorithms) is the downsampling of \textbf{HR} x3. *For $QMR_{512}$ images input is upscaled to 512x512 using a circular padding, thus, high-level features (from these QMRNets) activity are added in these columns.}
\label{fig:qmrimages1}
\end{figure}

\begin{figure}[H]
\centering
\begin{subfigure}[c]{\linewidth}
\small
\begin{tabular}{lllllllll}
    \textit{LR} & \hspace{18pt} vanilla & \hspace{8pt} $QMR_{\sigma,256}$ & \hspace{2pt} $QMR_{\sigma,512*}$ & \hspace{-6pt} $QMR_{rer,256}$  & \hspace{-6pt} $QMR_{rer,512*}$ &  \hspace{-6pt} $QMR_{snr,256}$ & \hspace{-6pt} $QMR_{snr,512*}$ & \hspace{8pt} \textbf{HR} 
\end{tabular}
\normalsize
\end{subfigure}
\begin{subfigure}[c]{\linewidth}
\includegraphics[width=\textwidth]{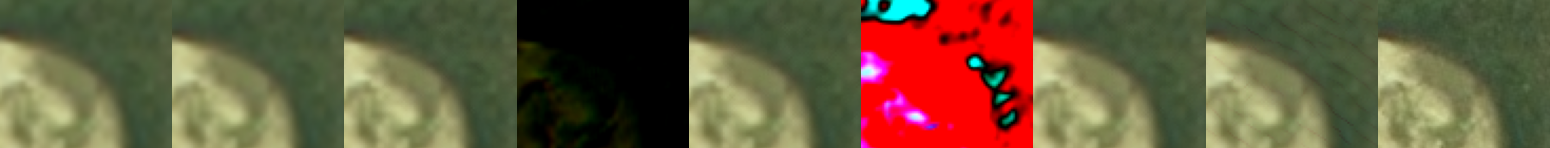}
\end{subfigure}
\begin{subfigure}[c]{\linewidth}
\includegraphics[width=\textwidth]{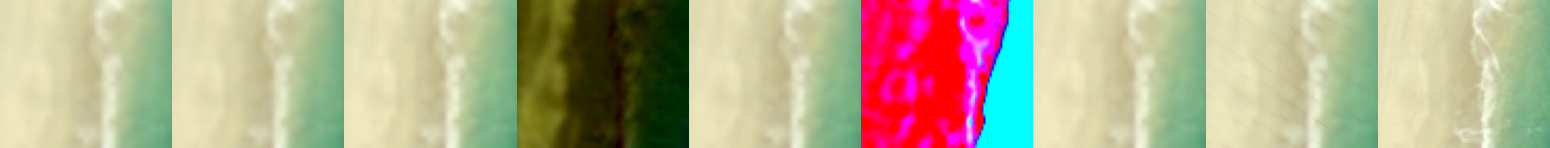}
\end{subfigure}
\begin{subfigure}[c]{\linewidth}
\includegraphics[width=\textwidth]{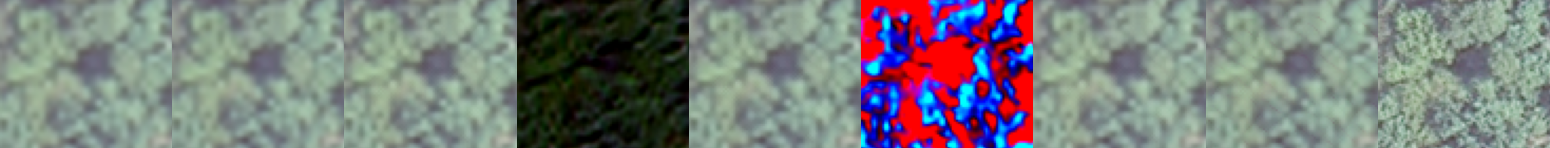}
\begin{subfigure}[c]{\linewidth}
\includegraphics[width=\textwidth]{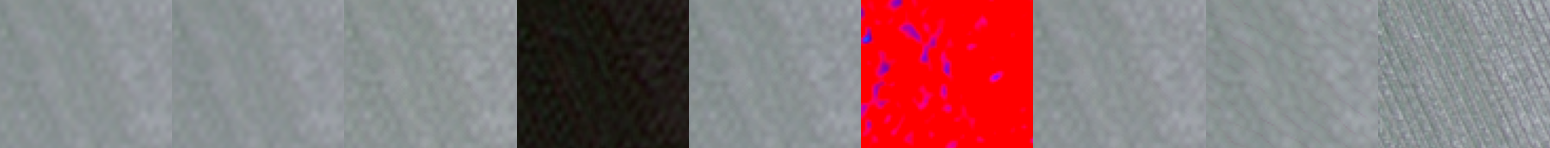}
\end{subfigure}
\end{subfigure}
\begin{subfigure}[c]{\linewidth}
\includegraphics[width=\textwidth]{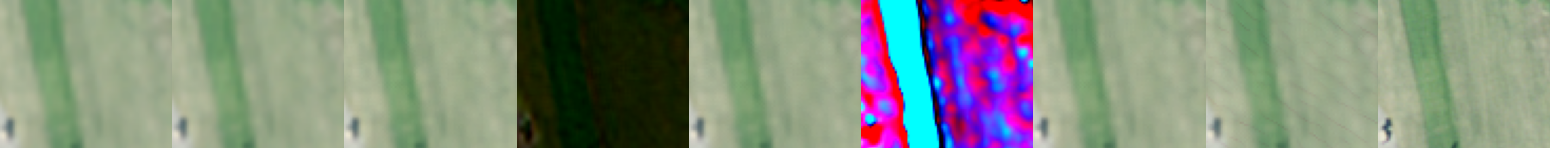}
\end{subfigure}
\begin{subfigure}[c]{\linewidth}
\includegraphics[width=\textwidth]{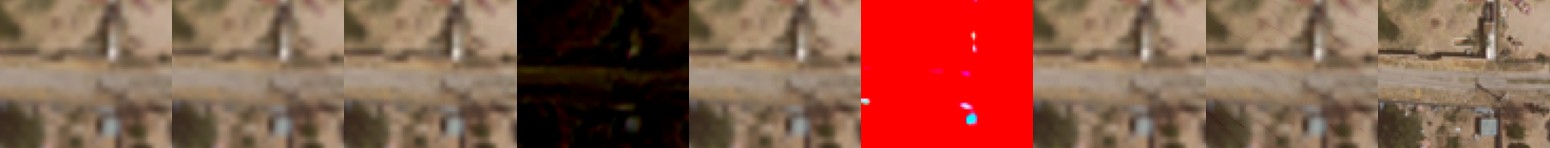}
\end{subfigure}
\begin{subfigure}[c]{\linewidth}
\includegraphics[width=\textwidth]{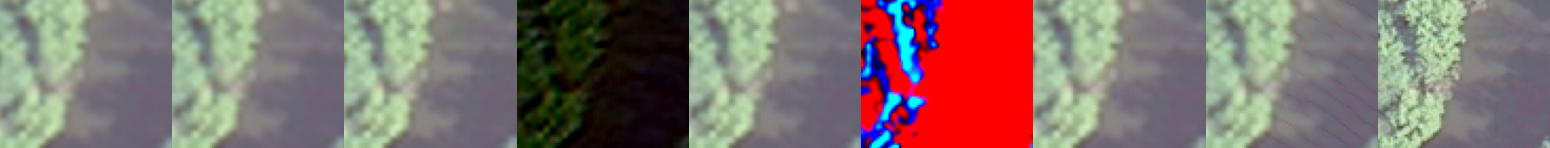}
\end{subfigure}
\begin{subfigure}[c]{\linewidth}
\includegraphics[width=\textwidth]{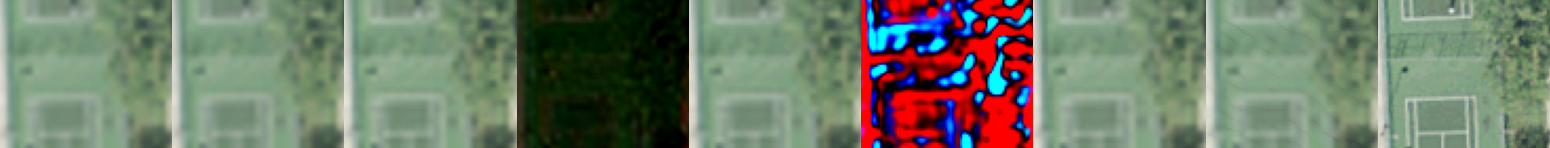}
\end{subfigure}
\begin{subfigure}[c]{\linewidth}
\includegraphics[width=\textwidth]{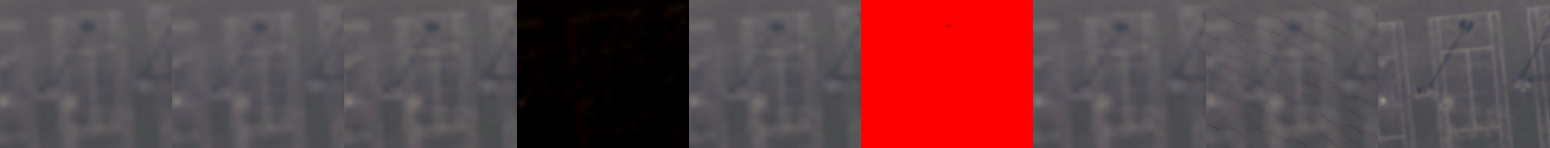}
\end{subfigure}
\begin{subfigure}[c]{\linewidth}
\includegraphics[width=\textwidth]{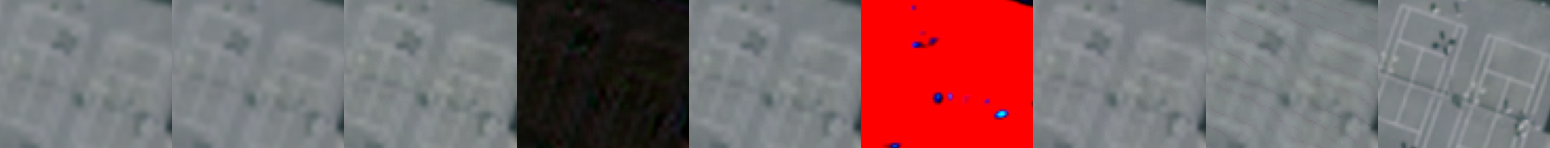}
\end{subfigure}
\caption{Examples (land and crops) of UCMerced images UCMerced images (crops of 256x256, with a zoom of x.40 i.e. 100x100) and each QMRNet Algorithm output (QMRLoss$_{L1}$). \textit{LR} (corresponding to input on algorithms) is the downsampling of \textbf{HR} x3. *For $QMR_{512}$ images input is upscaled to 512x512 using a circular padding, thus, high-level features (from these QMRNets) activity are added in these columns.}
\label{fig:qmrimages2}
\end{figure}

\end{document}